\documentclass[sigconf]{acmart}

\usepackage{booktabs} 
\setcopyright{rightsretained}
\usepackage{lipsum}
\usepackage{graphicx}
\usepackage{epsfig}
\usepackage{graphicx}
\usepackage{amsmath}
\usepackage{amssymb}
\usepackage{breqn}
\usepackage{mathtools}

\newenvironment{packed_enumerate}{
    \begin{enumerate}
        \setlength{\itemsep}{1pt}
        \setlength{\parskip}{0pt}
        \setlength{\parsep}{0pt}
    }{\end{enumerate}}
\newcommand{\ie}{\textit{i.e.,}}
\newcommand{\etal}{\textit{et al.}}
\newcommand{\eg}{\textit{e.g.,}}
\newcommand{\totaluser}{47}

\acmConference[ACMMM'18]{ACM Multimedia conference}{July 2018}{Seoul, Korea}
\acmYear{2018}
\copyrightyear{2018}

\acmArticle{4}
\acmPrice{15.00}

\acmSubmissionID{122}
\begin{document}
\title{Aesthetic-Driven Image Enhancement by Adversarial Learning}

%
%

%

\author{Yubin Deng}
\authornote{$\{dy015, ccloy, xtang\}$@ie.cuhk.edu.hk}
\affiliation{%
  \institution{CUHK-Sensetime Joint Laboratory, \ \ The Chinese University of Hong Kong}
}
\author{Chen Change Loy}
\affiliation{%
  \institution{CUHK-Sensetime Joint Laboratory, \ \ The Chinese University of Hong Kong}
}
\author{Xiaoou Tang}
\affiliation{%
  \institution{CUHK-Sensetime Joint Laboratory, \ \ The Chinese University of Hong Kong}
}

%
%
%
%

\begin{abstract}
We introduce EnhanceGAN, an adversarial learning based model that performs automatic image enhancement. 
Traditional image enhancement frameworks typically involve training models in a fully-supervised manner, which require expensive annotations in the form of aligned image pairs.  
In contrast to these approaches, our proposed EnhanceGAN only requires weak supervision (binary labels on image aesthetic quality) and is able to learn enhancement operators for the task of aesthetic-based image enhancement. 
In particular, we show the effectiveness of a piecewise color enhancement module trained with weak supervision, and extend the proposed EnhanceGAN framework to learning a deep filtering-based aesthetic enhancer.
The full differentiability of our image enhancement operators enables the training of EnhanceGAN in an end-to-end manner. 
We further demonstrate the capability of EnhanceGAN in learning aesthetic-based image cropping without any groundtruth cropping pairs.
Our weakly-supervised EnhanceGAN reports competitive quantitative results on aesthetic-based color enhancement as well as automatic image cropping, and a user study confirms that our image enhancement results are on par with or even preferred over professional enhancement.
\end{abstract}

%
%

\maketitle

\section{Introduction}
\label{sec:introduction}
\begin{figure}[t]
\begin{center}
\includegraphics[width=\linewidth]{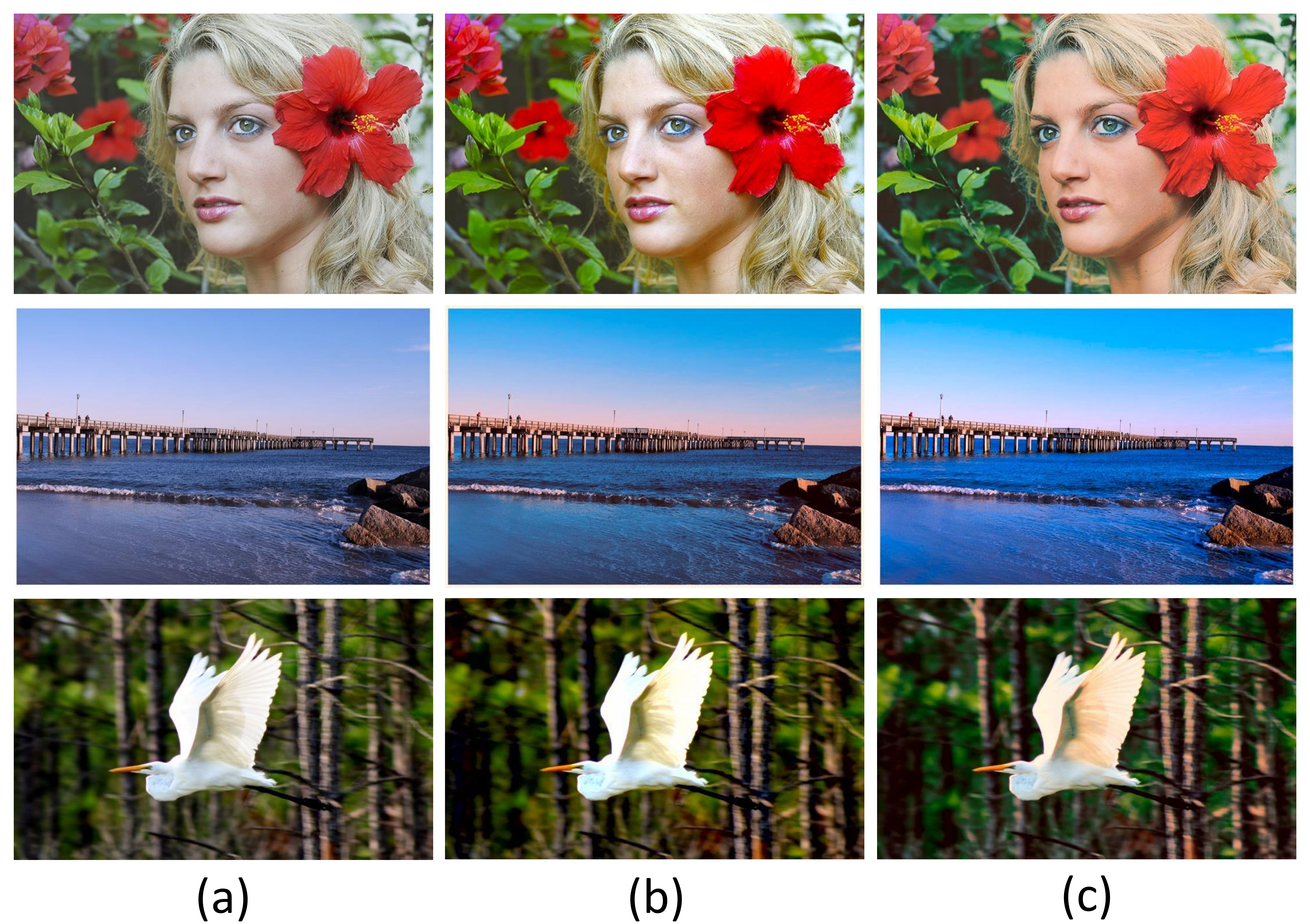}
\end{center}
\vskip -0.4cm
\caption[Caption for LOF]{Examples of image enhancement given original input (a). Can you distinguish which figure is enhanced by human and which by our adversarial learning model? (Answer in footnote\protect\footnotemark. Best viewed in color.)}\color{black}
\vskip -0.5cm
\label{fig:fig1}
\end{figure}

Image enhancement is considered a skillful artwork that involves transforming or altering a photograph using various methods and techniques to improve the aesthetics of a photo. Examples of enhancement include adjustments of color, contrast and white balance, as shown in Fig.~\ref{fig:fig1}. 
This task is conventionally conducted manually through some professional tools. Manual editing is time-consuming even for a professionally trained artist. 
While there are an increasing number of applications that allow casual users to choose a fixed set of filters and/or to alter the composition of a photo in more convenient ways, human involvement is still inevitable. 
Given the increasing amount of digital photographs captured daily with mobile devices, it is desirable to perform image enhancement with minor human involvement or in a smart and fully automatic manner.
\footnotetext{
\scriptsize
\raggedleft\rotatebox{180}{Row 3:\qquad$^{(b)}$ piecewise color enhancer \qquad $^{(c)}$ deep filtering\hfill}  \\
\raggedleft\rotatebox{180}{Row 2:\qquad$^{(b)}$ deep filtering \qquad $^{(c)}$ piecewise color enhancer\hfill}  \\ 
\raggedleft\rotatebox{180}{Row 1:\qquad$^{(b)}$ piecewise color enhancer \qquad $^{(c)}$ deep filtering\hfill} \\
\raggedleft\rotatebox{180}{\textit{Answer: None of them are the results from human editing}\hfill} }
Previous research efforts have shown some success in automating color enhancement~\cite{bhattacharya2010framework,lee2016automatic,sun2016photo,yan2014learning}, style transfer~\cite{gupta2017characterizing,huang2017arbitrary,johnson2016perceptual} and image cropping~\cite{chen2016automatic,chen2017quantitative,yan2013learning}. However, most of these models require full supervision. 
%
%
In particular, we need to provide input and manually-enhanced image pairs to learn the capability of color enhancement~\cite{yan2016automatic} or image re-composition~\cite{huang2015automatic,chen2016automatic,yan2013learning}. Unfortunately, such data is scarce due to the expensive cost of obtaining professional annotations. 
Ignatov et al.~\cite{ignatov2017dslr} has recently demonstrated the possibility of enhancing low-quality photos towards DSLR-quality. The training images, however, have to be captured by specialized time-synchronized hardware to form pairs, and further registered to remove misalignment between image pairs.
In this study, we present a novel approach that trains a weakly-supervised image enhancement model from unpaired images without strong human supervision.
%
In particular, we attempt to learn image enhancement from images with only binary labels on aesthetic quality, \ie, good or poor quality.
As the edited image should be closer (in the sense of aesthetic quality) to the photo collections by professional photographers compared to the original image with poor aesthetic quality, this notion can be well formulated in an adversarial learning framework~\cite{goodfellow2014generative}. 
Specifically, we have a discriminator $D$ that attempts to distinguish images of poor and good aesthetic quality. Such a network can be trained by an abundant amount of images from existing aesthetic datasets~\cite{murray2012ava,tang2013content}. A generator $G$, on the other hand,  generates a set of manipulation parameters given an image with poor aesthetic quality. The task of $G$ is to fool $D$ so that $D$ confuses the $G$'s outputs (\ie, enhanced images) as images with high quality.

\noindent The main contribution of this study is three-fold:
\begin{packed_enumerate}
  \item EnhanceGAN leverages abundant images that are annotated only with good and poor quality labels. \textit{No knowledge of the groundtruth enhancement action is given to the system}. Image enhancement operators are learned in a weakly-supervised manner through adversarial learning driven by aesthetic judgement.
  \item The framework permits multiple forms of color enhancement. We carefully design a piecewise color enhancement operator and a deep filtering-based enhancer to be fully-differentiable for end-to-end learning in an adversarial learning framework. 
  \item EnhanceGAN is extensible to include further image enhancement schemes (provided that the enhancement operations are fully-differentiable). We explore such capability of EnhanceGAN for aesthetic-based automatic image cropping and present competitive results. 
\end{packed_enumerate}
Owing to the subjective nature of image enhancement, we show the effectiveness of EnhanceGAN for image enhancement in two sets of evaluations. We quantitatively evaluate the performance of color enhancement by multiple state-of-the-art aesthetic evaluators on reserved unseen images and we show quantitative results for automatic image cropping on a standard benchmark dataset. We also perform a blind user study to compare our method with human enhancement.


\section{Related Work}
\label{sec:related_work}
\noindent\textbf{Aesthetic Quality Assessment.}
%
The task of aesthetic quality assessment is to distinguish high-quality photos from low-quality ones based on human perceived aesthetics. Previous successful attempts train convolutional neural networks for binary classification of image quality~\cite{lu2014rapid,ma2017lamp} or aesthetic score regression~\cite{kong2016photo}. We refer readers to a comprehensive study~\cite{deng2017image} on the state-of-the-art models on image aesthetic assessment. Although the focus of image enhancement is not on assessing the quality of a given image, our work is closely related to this research domain in the sense that image enhancement aims at improving the aesthetic quality of the given input. 

\begin{figure*}
\begin{center}
\includegraphics[width=0.95\linewidth]{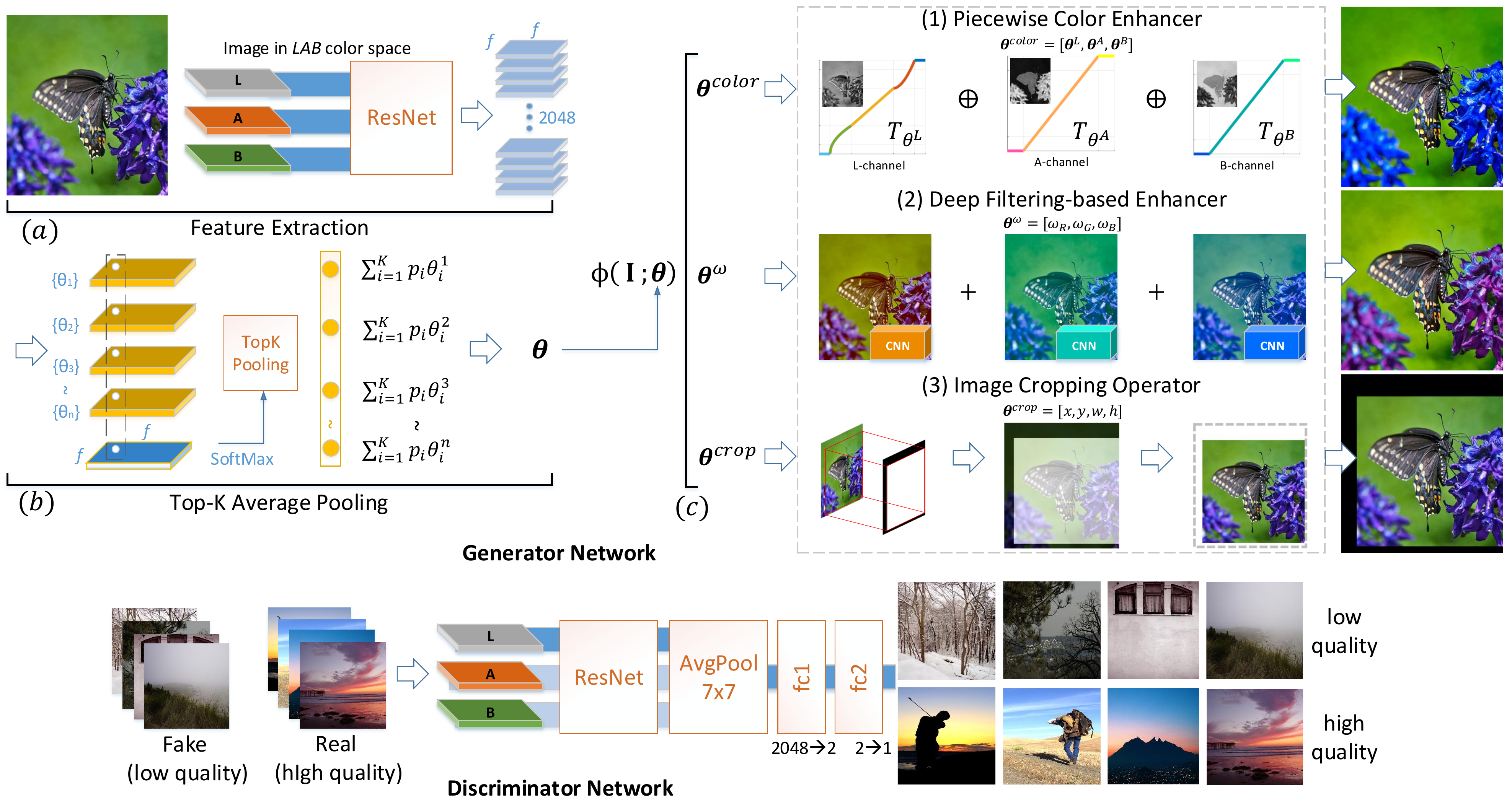}
\end{center}
\vskip -0.4cm
\caption{\textbf{The architecture of our proposed EnhanceGAN framework}. ResNet module is the feature extractor; in this work, we use the ResNet-101~\cite{he2016deep} and removed the last average pooling layer and the final fc layer. (Best viewed in color.)}
\label{fig:architecture}
\end{figure*}

\noindent\textbf{Automatic Image Enhancement:}
The majority of techniques for image manipulation with the goal to enhance the aesthetic quality of an image can be divided into two genres, namely (1) color enhancement and (2) image re-composition. Pixel-level manipulation and image restoration (\textit{e.g.,} super resolution~\cite{dong2016image}, de-haze~\cite{he2011single} and de-artifacts~\cite{wang2016d3}) are also closely related to image enhancement but is beyond the focus in this work.
\\
\noindent\textbf{Color Enhancement.}
The visual quality of an image can be enhanced by color adjustment, where regression models and ranking models have been trained to map the input image to a corresponding enhanced groundtruth~\cite{faridul2014survey}. Such color mappings are learned~\cite{yan2014learning} from a small set of labeled data by professional editors. To alleviate the lack of sufficient labeled data, recent research efforts formulate the color enhancement problem as the color transfer problem~\cite{hwang2014color,reinhard2001color}. In particular, the popular exemplar-based color transfer approaches~\cite{lee2016automatic,sun2016photo} seek to retrieve the most suitable matching exemplar based on image content and perform color transfer onto the given input. However, they suffer from potential visual artifacts due to erroneous exemplars. 
Ignatov \etal~\cite{ignatov2017dslr} present a fully-supervised approach aided by adversarial learning, where they seek to improve low-quality images towards DSLR quality. However, their method still requires carefully aligned input and groundtruth pairs and the effectiveness of their model is limited to images captured by particular mobile devices.  
Stylistic image enhancement~\cite{yan2016automatic} and creative style transfer~\cite{gupta2017characterizing,huang2017arbitrary,johnson2016perceptual,ulyanov2016instance} are also closely related to color enhancement, but their focus is to transform an input image into an output that matches the artistic style of a given exemplar, instead of focusing on improving the aesthetic quality of the image. 
Aesthetic-oriented evaluation typically does not apply on those methods. 
Unlike style transfer studies~\cite{yan2016automatic,gupta2017characterizing}, our weakly supervised model is able to enhance an image through chrominance and even spatial manipulations driven by content and aesthetic quality of the image. Thus each individual image would experience different manipulations. Unlike exemplar matching~\cite{lee2016automatic,sun2016photo}, our framework does not require finding the correct exemplars for color/style transfer, and hence our model is not limited by the subset of exemplars.
\\
\noindent\textbf{Cropping and Re-targeting.} 
Image cropping and re-targeting aim at finding the most visually significant region based on aesthetic value or human attention focus. Aesthetic-based approaches~\cite{chen2016automatic,islam2016survey,yan2013learning,chen2017learning} evaluate the crop window candidates based on handcrafted low-level features or learned aesthetic features, while attention/composition-based approaches~\cite{fang2014automatic,choi2016object,huang2015automatic,jaiswal2015saliency} rely on image saliency and produce the cropping window encapsulating the most salient region. 
These systems for cropping and re-targeting are mostly based upon a limited amount of labeled cropping data ($\sim$1000 training image pairs in total), where the cropping problem is modeled as window regression or window candidate classification in a fully-supervised learning manner~\cite{chen2016automatic,chen2017quantitative}. Network fine-tuning with pre-trained convolutional neural network has obtained some success with extensive data augmentation~\cite{deng2017image}.
\color{black}
The non-parametric images/windows retrieval-based method by Chen \etal~\cite{chen2017learning} also features a weakly-supervised model for image cropping. 
\color{black}
In this work, we present an alternative learning-based weakly-supervised attempt and demonstrate the capability of the proposed EnhanceGAN framework in extending the image enhancement tasks to automatic image cropping, and show competitive results on a standard benchmark dataset. 
\\


\section{Aesthetic-Driven Image Enhancement}
\label{sec:methodology}
We formulate the problem of image enhancement in an adversarial learning framework~\cite{goodfellow2014generative}. Specifically, our proposed approach builds upon Wasserstein GAN (W-GAN) by Arjovsky \etal~\cite{arjovsky2017wasserstein}. The full architecture of our proposed framework is shown in Fig.~\ref{fig:architecture}. 

\subsection{Preliminary}
Generative Adversarial Network (GAN)~\cite{goodfellow2014generative} has shown a powerful capability of generating realistic natural images. Typical GANs contain a generator \textit{G} and a discriminator \textit{D}, and it was proven~\cite{goodfellow2014generative} that the minimax game
\begin{equation}
\begin{split}
\min_G \max_D
 V(D,G) = \mathbb{E}_{\mathbf{I} \sim p_{data}}[\text{log}D(\mathbf{I})] \\+ \mathbb{E}_{\mathbf{z} \sim p_{\mathbf{z}}}[\text{log}(1-D(G(\mathbf{z})))]
\end{split}
\end{equation}
would reach a global optimum when $p_g$ converges to the real data distribution $p_{data}$, where $p_g$ is the distribution of the samples $G(\mathbf{z})$ obtained when $\mathbf{z} \sim p_{\mathbf{z}}$, and $\mathbf{z}$ is a random or encoded vector. 
In this work we follow the practice in~\cite{arjovsky2017wasserstein} and adopt the loss function based on Wasserstein distance,
\begin{dmath}
L = \mathbb{E}_{\mathbf{I} \sim p_{data}}[f_{W}(\mathbf{I})] - \mathbb{E}_{\mathbf{I} \sim p_{gen}}[f_{W}(\mathbf{I})],
\label{eq:w_loss}
\end{dmath}
where $f_W(\cdot)$ is a $K$-Lipschitz function parameterized by $W$, which is approximated by our discriminator network \textit{D} as detailed in Sec.~\ref{sec:discriminator_network}.

\subsection{Generator Network (Net-\textit{G})}
Different from most existing GAN frameworks, our generator does not generate images by itself. 
Instead, the generator \textit{G} in our EnhanceGAN is responsible for learning the image enhancement operator ${\Phi}(\cdot)$, according to which the input image will be transformed to the enhanced output: 
\begin{equation}
    \mathbf{I}^{\text{output}} =\Phi(\mathbf{I}; \boldsymbol{\theta}) = T_{\boldsymbol{\theta}}(\mathbf{I}),
\end{equation} 
where $T$ denotes the fully-differentiable transformation applied to the input image parameterized by $\boldsymbol{\theta} = G(\mathbf{I})$.  
The base architecture of our generator network is a ResNet-101~\cite{he2016deep} without the last fully-connected layer, and we further remove the last pooling layer to preserve spatial information in the feature maps. As such, this ResNet module acts as a fully-convolutional feature extractor given an input image (see Fig.~\ref{fig:architecture}$a$). 
The 2048 output feature maps produced by the ResNet module has a spatial size $f \times f$ and is subsequently utilized in our enhancement parameter generation modules. 
In this work, we explore multiple forms of image aesthetic enhancement, including two fully-differentiable color enhancement operators and an image editing operator for automatic image cropping. 
%

\subsubsection{Piecewise Color Enhancer}
\label{subsubsec:piecewise_color_enhancer}
Image brightness and lighting contrast can be adjusted based on the luminance channel of the  image in the CIELab color space, whereas image chrominance resides in the other two channels. Hence, the color enhancement operator $\Phi(\mathbf{I})$ can be applied to an image in a piecewise manner. To this end, the piecewise color enhancement module is designed to learn a set of parameters
$\boldsymbol{\theta}^{L}, \boldsymbol{\theta}^{AB} \in \{\boldsymbol{\theta}^{color} | \boldsymbol{\theta}^{color} = G(\mathbf{I})\}$,
where $\boldsymbol{\theta}^{L}$ and $\boldsymbol{\theta}^{AB}$ denotes respectively the adjustments for better lighting and chrominance. Specifically, we follow the idea behind gamma correction~\cite{pelli1991accurate} to adjust the brightness and contrast of an image in pixel level (\ie, the L channel of image $\mathbf{I}$ in the CIELab color space: $I^L$) by designing a piecewise transformation function $T_{\boldsymbol{\theta}^{L}}$ (see Fig.~\ref{fig:architecture}$c$) defined on each pixel $m \in I^L$:
\[
\begin{array}{@{} r @{} c @{} l @{} }
&T_{\boldsymbol{\theta}^{L}}(m) &{}=\displaystyle
\begin{cases}
0 &\text{if } m \leq b  \\
k_{1} (m-b)^{\frac{1}{p}} &\text{if } b < m \leq a \\
m &\text{if } a < m \leq 1-a \\
k_{2} (m-k_{3})^{\frac{1}{q}} + k_{3}& \text{if } 1-a < m \leq 1-b \\ 
1 &\text{if } m \geq 1-b
\end{cases},
\end{array}
\]
where $k_1 =a(a - b)^{-\frac{1}{p}}$, $k_2 =a(a - b)^{-\frac{1}{q}}$ and $k_3 = 1-a$ to ensure that $T_{\boldsymbol{\theta}^{L}}$ is continuous. We further constrain $p \geq 1$ in order to lighten the dark regions and $0 < q < 1$ to darken the over-exposed region. Similarly, we follow ``The LAB Color Move''\footnote{The LAB Color Move: \url{https://goo.gl/i2ppcw}} and the curve adjustment instructed in~\cite{hosie2011new} and design a similar process to enhance the image color. In particular, the adjustments $T_{A}$ and $T_{B}$ defined respectively on pixels $m \in I^A$ and $m \in I^B$ (\ie, the A and B channels in image $\mathbf{I}$, see Fig.~\ref{fig:architecture}$c$) can be formulated as follows:
\[
\begin{array}{@{} r @{} c @{} l @{} }
&T_{\boldsymbol{\theta}^{A}}(m) &{}=\displaystyle
\begin{cases}
0 &\text{if } m \leq \alpha  \\
\frac{1}{1 - 2\alpha}(m-\alpha) &\text{if } \alpha < m < 1-\alpha \\
1 &\text{if } m \geq 1-\alpha
\end{cases},
\end{array}
\]
\[
\begin{array}{@{} r @{} c @{} l @{} }
&T_{\boldsymbol{\theta}^{B}}(m) &{}=\displaystyle
\begin{cases}
0 &\text{if } m \leq \beta  \\
\frac{1}{1 - 2\beta}(m-\beta) &\text{if } \beta < m < 1-\beta \\
1 &\text{if } m \geq 1-\beta
\end{cases}.
\end{array}
\]
Under this formulation, the parameter sets $\boldsymbol{\theta}^{L}=[a,b,p,q]$ and $\boldsymbol{\theta}^{AB} = [\boldsymbol{\theta}^{A}, \boldsymbol{\theta}^{B}]=[\alpha, \beta]$ can be learned end-to-end. Our piecewise color enhancement operator can therefore be written as
\begin{equation}
\label{eq:piece_wise_eq}
\begin{split}
\Phi(\mathbf{I}; \boldsymbol{\theta}) =& \ T_{\boldsymbol{\theta}^{color}}(\mathbf{I}) \\=& \ T_{\boldsymbol{\theta}^{L}}(I^L) \oplus T_{\boldsymbol{\theta}^{A}}(I^A) \oplus T_{\boldsymbol{\theta}^{B}}(I^B),
\end{split}
\end{equation} 
where $\oplus$ denotes channel-wise concatenation in the CIELab color space. Directly learning one single set of such parameters may not be optimal as color enhancement prediction is multi-modal to some extents -- an image can have several plausible color enhancement solutions (\eg, an enhancement solution can simply adjust the color saturation of an image, or further tuning lighting contrast from low to high or vice versa). 

With inspirations drawn from attention models~\cite{xu2015show}, our piecewise enhancement operator is built by appending a convolution layer ($2048 \rightarrow  7$) with kernel size $1 \times 1$ to the ResNet module.
In particular, the first 6 feature maps correspond to the candidate sets of the color adjustment parameters $[a,b,p,q,\alpha, \beta]$.
The 7$^{th}$ feature map is an $f \times f$ softmax probability map corresponding to $f^{2}$ possible predictions $\boldsymbol{\theta}_{i}^{color} = [a_i, b_i, p_i, q_i,\alpha_i, \beta_i], i \in \{1, 2, ..., f^{2}\}$, where $prob(\boldsymbol{\theta}^{color} = \boldsymbol{\theta}^{color}_{i} | \mathbf{I}) = p_i$. Top-K average pooling~\cite{wang2017untrimmednets} is adopted to aggregate the parameter candidates with the highest probabilities (see Fig.~\ref{fig:architecture}$b$). 
We show in the experiment section that our piecewise color enhancement operator is able to improve both the color and lighting contrast of the image as compared with the input (see Fig.~\ref{fig:fig1}, \ref{fig:architecture}).

\subsubsection{Deep Filtering-based Enhancer}
\label{subsubsec:deep_filtering}
The piecewise color enhancement operator is limited to learning enhancement parameter $\boldsymbol{\theta}^{color} = [\boldsymbol{\theta}^{L}, \boldsymbol{\theta}^{A}, \boldsymbol{\theta}^{B}]$ for a pre-defined set of transformations $\phi_{1}(\cdot) =T_{\boldsymbol{\theta}^{L}},\phi_{2}(\cdot)=T_{\boldsymbol{\theta}^{A}}$ and $\phi_{3}(\cdot)= T_{\boldsymbol{\theta}^{B}}$. We can extend equation~\eqref{eq:piece_wise_eq} to a general form by writing the enhancement operator $\Phi(\cdot)$ to be a linear combination of individual transforming operations $\phi_{1}(\cdot), \phi_{2}(\cdot), ..., \phi_{n}(\cdot)$ as follows,
\begin{equation}
\Phi(\mathbf{I}; \boldsymbol{\theta})  = \Phi(\mathbf{I}; \boldsymbol{\theta}^{\omega})=\sum\nolimits_{i=1}^{n}\omega_{i}\phi_{i}(\mathbf{I})
\end{equation}
\begin{equation}
\boldsymbol{\theta}^{\omega}=[\omega_1, \omega_2, ..., \omega_n] = G(\mathbf{I})
\end{equation}
In particular, $\phi_{i}$ can be any form of image enhancement transformation provided that it is fully-differentiable. Under this formulation, the weight parameters $\boldsymbol{\theta}^{\omega}$ can be learned end-to-end by a convolution layer ($2048 \rightarrow  n+1$) where the $(n+1)^{th}$ feature map is also a $f \times f$ softmax probability map for Top-K average pooling similar to that of the piecewise color enhancement operator (see Sec.~\ref{subsubsec:piecewise_color_enhancer}). In this work, we choose three default 
$R$,$G$,$B$ color enhancement filters from \textit{Adobe Photoshop}\footnote{The three filters are: the \textit{cooling80} filter, the \textit{waming85} filter and the \textit{underwater} filter. The three filters represent respectively blue-ish (\#$00b5ff$), red-ish (\#$ec8a00$) and green-ish (\#$00c2b1$) effect under default settings.} and approximate each of the filtering operations with a 3-layer convolutional neural network (see Fig.~\ref{fig:architecture}$c$). 
More color filtering operators $\phi(\cdot)$ with learnable parameter $\boldsymbol{\theta}$ can be easily extended in this manner provided that the transformations  are differentiable. We show in the experiment section that the deep filtering-based aesthetic enhancer produce smooth and harmonious color improvement as compared to the piecewise color enhancer (see Fig.~\ref{fig:BigFig}, \ref{fig:linzhe}, \ref{fig:visual_results}).

\subsubsection{Image Cropping Operator}
We further extend the image enhancement operation to explore the possibility of learning the task of aesthetic-based image cropping without any cropping labels. The goal of image cropping is to produce a set of cropping coordinates $\boldsymbol{\theta}^{crop} = [x, y, w, h]$ given an input image. This can be achieved by a convolution layer ($2048 \rightarrow  (4+1)$) where the 5$^{th}$ feature map is the softmax probability map for Top-K average pooling. It is worth mentioning that directly extracting a sub-image from the input given the learned cropping coordinates $\boldsymbol{\theta}^{crop}$  is not differentiable. To ensure that the gradients can back-propagate to the cropping parameters via the transformation $\Phi(\mathbf{I}) = T_{\boldsymbol{\theta}^{crop}}(\mathbf{I})$, bilinear sampling from a sampling grid~\cite{jaderberg2015spatial} is used to sub-sample the input image based on coordinates output $\boldsymbol{\theta}^{crop}$ of the cropping module (see Fig.~\ref{fig:architecture}$c$). 




\subsubsection{Generator Loss Function $L_{G}$}
\label{sec:generator_loss}
We formulate the loss function for the generator network as a weighted sum of an adversarial loss component $L_{gan}$ as well as the regularization component $L_{reg}$. These terms are weighted to ensure that the loss terms are balanced in their scales. This formulation makes the training process more stable and has better performance (see Sec.~\ref{sec:quantitative}).

\noindent\textbf{Adversarial Loss:}
Following Arjovsky \etal~\cite{arjovsky2017wasserstein}, the adversarial gradient to the generator network \textit{G} is computed from the following loss function:
\begin{equation}
L_{gan} = \frac{1}{n} \sum\nolimits_{i=1}^{n}f_{W}(\mathbf{I}^{\text{output}}_{i}).
\label{eq:gan_loss}
\end{equation}

\noindent\textbf{Regularization Loss:}
As regularization, we use the feature reconstruction loss~\cite{johnson2016perceptual} to account for the semantic difference between the enhanced/cropped image and the input as measured by feature similarity:
\begin{equation}
L_{reg^{1} } = \frac{1}{n}\sum\nolimits_{i=1}^{n}|| f_{vgg}(\mathbf{I}^{\text{output}}_{i}) - f_{vgg}(\mathbf{I}_{i}) ||_{2}^{2},
\label{eq:per_loss}
\end{equation}
where $f_{vgg}$ denotes the $fc7$ feature output from the VGG-16 network~\cite{simonyan2014very} trained for ImageNet. Also, the notion that an edited image should have better aesthetic quality (lower $f_{W}(\cdot)$ values) than the original further gives us an intuitive loss for further regularizing the end-to-end training:
\begin{equation}
L_{reg^{2}} = \frac{1}{n}\sum\nolimits_{i=1}^{n} \phi(\mathbf{I}^{\text{output}}_{i}, \mathbf{I}_{i})
\label{eq:reg_loss}
\end{equation}
where $
   \phi(\mathbf{I}^{\text{output}}_{i}, \mathbf{I}_{i}) \\=%
   \begin{cases}
     0 & \text{if $ f_{W}(\mathbf{I}^{\text{output}}_{i}) < f_{W}(\mathbf{I}_{i}) $} \\
     || f_{W}(\mathbf{I}^{\text{output}}_{i}) - f_{W}(\mathbf{I}_{i}) ||_{2}^{2} \ & \text{otherwise}.
   \end{cases} $
\vskip 0.2cm
\noindent We show in the experiment section that the regularization facilitates the aesthetic-driven adversarial learning of the generator network (see Table~\ref{tab:quantitative_aesthetics1}, \ref{tab:quantitative_aesthetics2}).


\subsection{Discriminator Network (Net-\textit{D})}
\label{sec:discriminator_network}
The proposed framework consists of a discriminator network that is able to assess image aesthetic quality. The discriminator network \textit{D} is designed to share the ResNet-101~\cite{he2016deep} base architecture of \textit{G} during pre-training. As shown in Fig.~\ref{fig:architecture}, the last layer for 1000-class classification in the original ResNet-101 is replaced by a 2-neuron fully-connected layer. We pre-train discriminator \textit{D} for binary aesthetic classification with the cross-entropy loss as in~\cite{deng2017image}.
After pre-training, the discriminator network \textit{D} is appended with another 1-neuron fully-connected layer to perform output aggregation as an approximator to $f_W$ in Eq.~(\ref{eq:w_loss}, \ref{eq:gan_loss}, \ref{eq:per_loss}, \ref{eq:reg_loss}). Deriving from Eq.~\ref{eq:w_loss}, the loss function $L_{D}$ in subsequent adversarial training can be written as:
\begin{equation}
L_{D} = E_{\mathbf{I} \sim p_{\text{good}}}[f_{W}(\mathbf{I}^{\text{good}})] - E_{\boldsymbol{\theta} \sim \mathbf{I}^{\text{bad}}}[f_{W}(\mathbf{I}^{\text{output}})],
\label{eq:L-D}
\end{equation}
where $\mathbf{I}^{\text{output}} = T_{\{\boldsymbol{\theta}\}}(\mathbf{I}^{\text{bad}}), \ \mathbf{I}^{\text{bad}} \sim p_{\text{bad}}$.

\begin{figure}[t]
\begin{center}
\includegraphics[width=\linewidth]{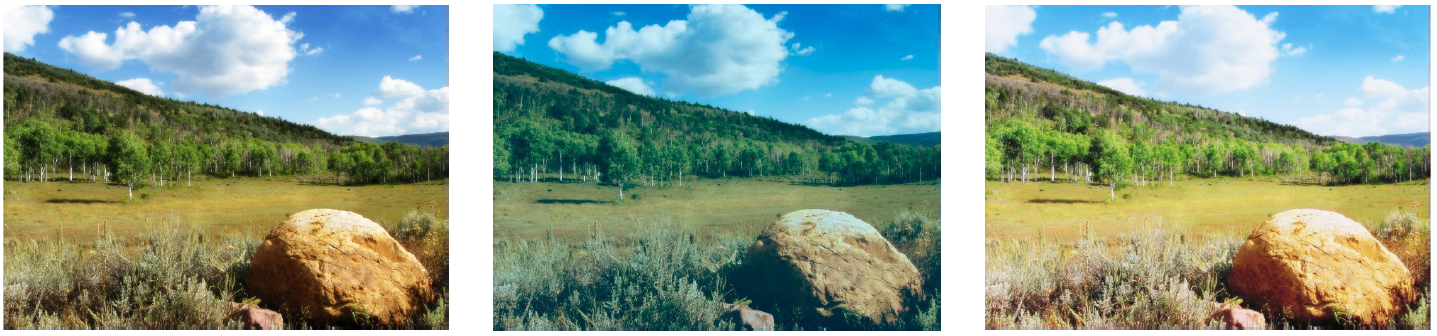}
\includegraphics[width=\linewidth]{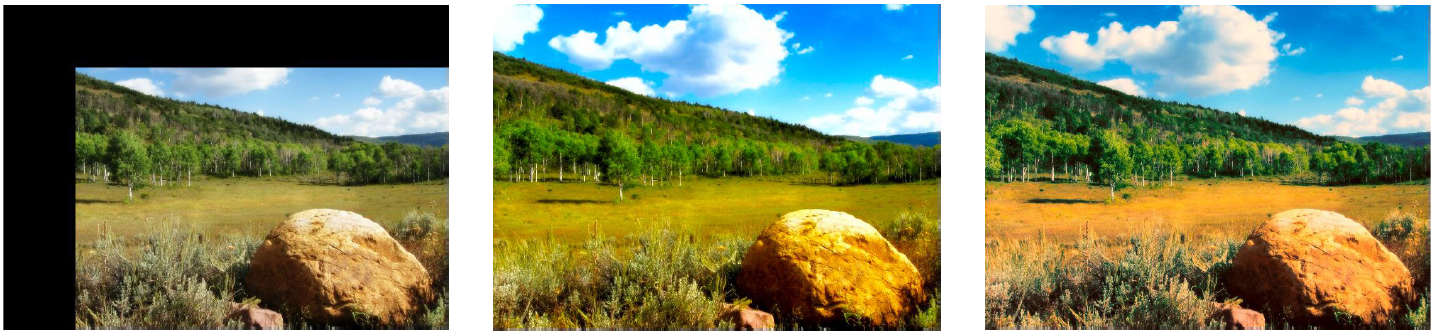}
\end{center}
\vskip -0.3cm
\caption{From left to right, top to bottom: (1) input image (2) random combination of pre-trained filters (3) DSLR$_{iphone}$~\cite{ignatov2017dslr} (4) EnhanceGAN: image cropping (5) EnhanceGAN: piecewise color enhancer (5) EnhanceGAN: deep filtering-based enhancer. (Best viewed in color. More results in the supplementary material.) }
\label{fig:BigFig}
\end{figure}
\begin{table}[h]
\centering
\caption{Quantitative evaluation for color enhancement on the unseen $Val^{100}$ images by multiple aesthetic evaluators. \textit{AVA-net} and \textit{CUHK-net} denotes the ResNet-based evaluators finetuned for binary image aesthetic assessment with AVA dataset and CUHK-PQ dataset, respectively (see Sec.~\ref{sec:dataset}). The averaged softmax scores for all images in $Val^{100}$ are shown as final results.
Our weakly supervised EnhanceGAN is reasonably competitive for the task of aesthetic-based color enhancement, as is also validated by additional state-of-the-arts image aesthetic assessment models~\cite{kong2016photo,deng2017image}.
}
\setlength{\tabcolsep}{1.9 pt}
\begin{tabular}{lcccc}
Methods                                               & \textit{AVA-net} & \textit{CUHK-net} & RANK~\cite{kong2016photo} & DAN~\cite{deng2017image} \\ \hline 
Original Input                                                & 0.705     & 0.542       & 0.487       & 0.479                                    \\
DSLR$_{iphone}$~\cite{ignatov2017dslr}         & 0.624     & 0.449      & 0.476        & 0.359                                \\
DSLR$_{sony}$~\cite{ignatov2017dslr}              & 0.524     & 0.385      & 0.475        & 0.295                                    \\
DSLR$_{blackberry}$~\cite{ignatov2017dslr}      & 0.616     & 0.462      & 0.477        & 0.395                                   \\
Photoshop-Auto                                             & 0.687        & 0.530         & 0.481          & 0.447                                     \\
                                                     &            &             &               &                                                  \\                                                   
Piecewise Enhancer                                                     &            &             &               &                                                 \\ \hline
w/o $L_{gan}$                                  & 0.708     & 0.705      & 0.484        & 0.491                                         \\
w/o $L_{reg^{1}}, L_{reg^{2}}$                       & 0.751    &  0.755      & 0.492        & 0.527                                       \\
EnhanceGAN                                                 & 0.764     & $\mathbf{0.805}$      & 0.494      & 0.573                                  \\    
                                                     &            &             &               &                                                  \\
Deep Filtering                                                     &            &             &               &                                                 \\ \hline
random weights                               & 0.687     & 0.432      & 0.479          & 0.438                                        \\
EnhanceGAN                                                & $\mathbf{0.769}$     & 0.752      & $\mathbf{0.498}$          & $\mathbf{0.624}$   \\               
\end{tabular}
\label{tab:quantitative_aesthetics1}
\vskip -0.4cm
\end{table}
\section{Experiments}
Our weakly-supervised EnhanceGAN is tasked to learn image enhancement operators based only on binary labels on image aesthetic quality. Specifically, we train the EnhanceGAN with the benchmark datasets used in aesthetic quality assessment and perform quantitative evaluations on reserved unseen data. A user study is also performed to confirm the validity of our quantitative evaluation.

\subsection{Experimental Settings}
\label{sec:dataset}
\noindent\textbf{CUHK-PhotoQuality Dataset (CUHK-PQ)}~\cite{tang2013content}: The dataset contains 4,072 high-quality images and 11,812 low-quality images. This dataset is used to pre-train the feature extractor (ResNet module) of our EnhanceGAN, with $10\%$ of the images reserved for validation. We follow the training protocol as in~\cite{deng2017image} and pre-train the ResNet-module for binary image aesthetic assessment (see Sec.~\ref{sec:discriminator_network}), obtaining a balanced accuracy~\cite{deng2017image} of $94.3\%$ on the validation set. This pre-trained network (denoted as CUHK-Net) is also used as one of the quantitative aesthetic evaluators (see Table~\ref{tab:quantitative_aesthetics1},\ref{tab:quantitative_aesthetics2}).

\noindent\textbf{AVA Dataset}~\cite{murray2012ava}:
The Aesthetic Visual Analysis (AVA) dataset is by far the largest benchmark for image aesthetic assessment. Each of the 255,530 images is labeled with aesthetic scores ranging from 1 to 10. We follow~\cite{murray2012ava} and partition the images into high-quality set and low-quality set based on the average scoring. In this study we select a subset of low-quality images based on the semantic tags provided in the AVA data for analysis\footnote{This corresponds to nine classes in the AVA dataset, \ie, Landscape, Seascape, Cityscape, Rural, Sky, Water, Nature, Animals and Portraiture.}, covering a diverse set of images that require different enhancement operations for quality enhancement. In particular, the ``Real'' input to the discriminator in our EnhanceGAN is chosen from the top $30\%$ of the high-quality images. The ``Fake'' inputs are the low-quality images that have an average score $<5$. A total of $20,000$ low-quality images are used for training the EnhanceGAN, each paired with k=5 high-quality images sampled from the k-nearest neighbor (k-NN) in the feature space of $f_{vgg}$ (All training images are from the AVA standard training partition). \textit{The binary good/bad quality is the only form of supervision in our training framework. No groundtruth enhancement operation is provided}. 

\noindent\textbf{Test Data 1}:
We keep 100 random images from the standard test partition of AVA (denoted as ${Val}^{100}$) for evaluation.

\noindent\textbf{Test Data 2}:
The \textbf{MIT-Adobe FiveK Dataset}~\cite{bychkovsky2011learning} contains 5,000 images, each is enhanced by 5 experts towards personal quality improvement. The standard test partition of the FiveK dataset as in~\cite{bychkovsky2011learning} is used in for evaluation.
\begin{figure}[t]
\begin{center}
\includegraphics[width=\linewidth]{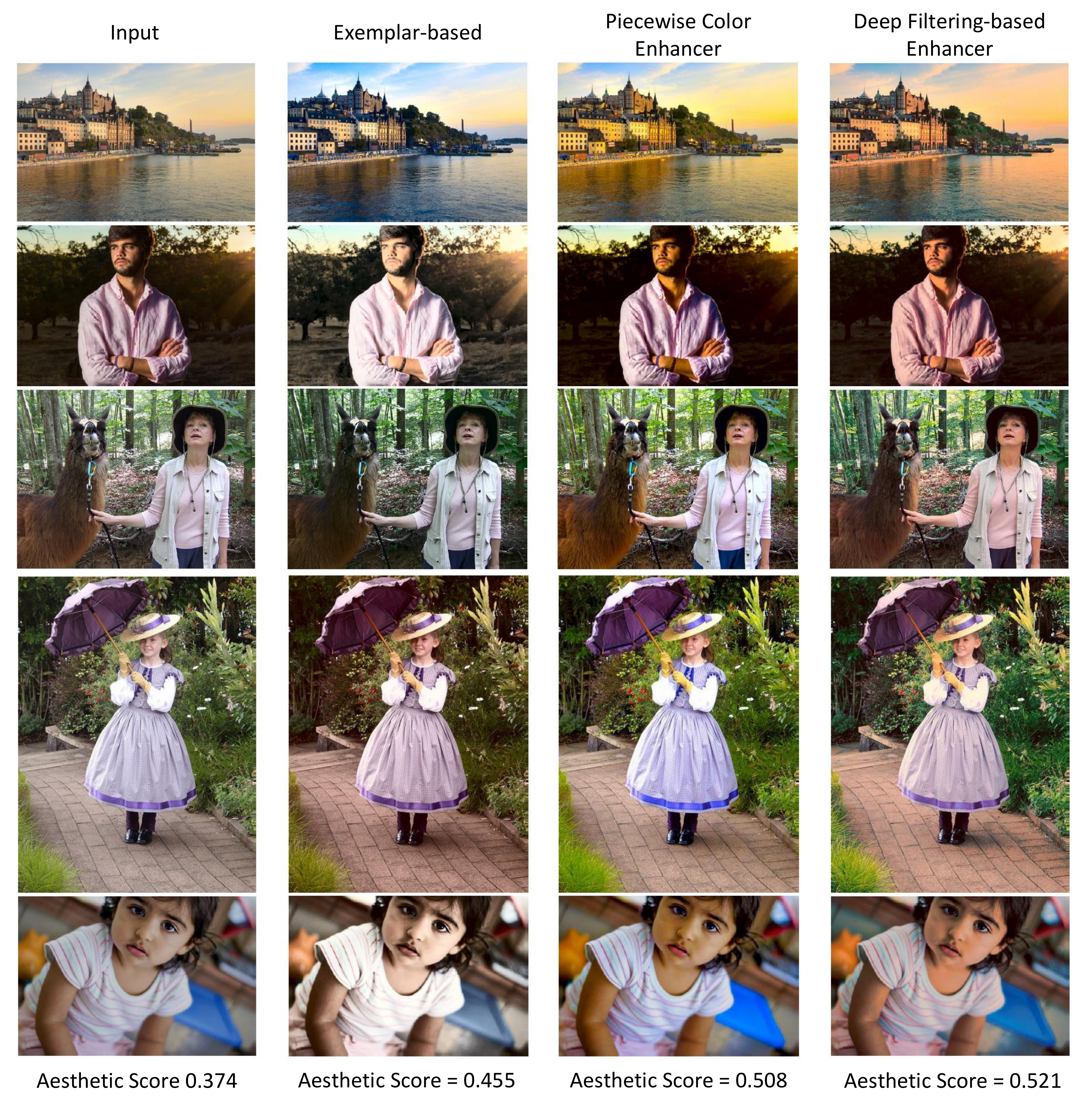}
\end{center}
\vskip -0.3cm
\caption{Visual results on the 5-image evaluation set in~\cite{lee2016automatic}. Column 1: input image; Column 2: output by the state-of-the-art exemplar-based method in Lee \etal~\cite{lee2016automatic}; Column 3: output by the piecewise color enhancer. Column 4: output by the deep filtering-based enhancer. The average aesthetic scores as evaluated by~\cite{deng2017image} are shown below. (Best viewed in color. More results in the supplementary material.)}
\label{fig:linzhe}
\end{figure}
\\
\noindent\textbf{Implementation Details}:
\noindent The generator network \textit{G} in our EnhanceGAN is fully convolutional, allowing for arbitrary-sized input in CIELab color space. We set the image input size to be $224\times224$, resulting in 2048 feature maps with spatial size of $7\times7$ (see Fig.~\ref{fig:architecture}$a$). We found K = 3 in Top-K averaging pooling (Fig.~\ref{fig:architecture}$b$) to be robust in producing parameter candidates. Using a larger K value or global average pooling (\textit{i.e.,} selecting from a large number of parameter candidates) could potentially suffer from noisy parameter predictions, while max pooling only concerns one single prediction and is prone to error. We use RMSprop~\cite{tieleman2012lecture} with a learning rate $lr_{G} = 5e^{-5}$ for the generator network and $lr_{D} = 5e^{-7}$ for the discriminator network after pre-training. A batch size of 64 is used to train each of the image enhancement operators.

\begin{table}[t]
\centering
\caption{\color{black}Quantitative evaluation for color enhancement on the MIT-Adobe FiveK Dataset. Top1-Expert (Top2-Expert) is the best (second best) scores among five groundtruth images produced by the experts as in the dataset. Our weakly-supervised EnhanceGAN is reasonably competitive as we have not used any expert labels in this dataset.\color{black}
}
\setlength{\tabcolsep}{2.8pt}
\begin{tabular}{lcccc}
Methods                                               & \textit{AVA-net} & \textit{CUHK-net} & RANK~\cite{kong2016photo} & DAN~\cite{deng2017image} \\ \hline 
Original Input                                                & 0.651     &  0.247       & 0.467       & 0.444                                    \\
Photoshop-Auto                                             & 0.638        & 0.277         & 0.397          & 0.319                                     \\                                             
Top1-Expert                                                  & 0.730        & 0.397         & 0.482          & 0.532                                     \\
Top2-Expert                                                 & 0.675        & 0.319         & 0.471          & 0.480                                     \\
                                             &     &          &        &                                     \\
EnhanceGAN                                                     &            &             &               &                                                 \\ \hline
Piecewise                                 & $\mathbf{0.731}$     & $\mathbf{0.516}$      & 0.476      & $\mathbf{0.502}$                                  \\    
Deep Filtering                               & 0.728     & 0.433      & $\mathbf{0.477}$         & 0.453   \\          
\end{tabular}
\label{tab:quantitative_aesthetics2}
\vskip -0.6cm
\end{table}

\subsection{Evaluations}
\label{sec:quantitative}
\noindent\textbf{Image Aesthetic Assessment:}
Evaluating aesthetic quality of enhanced images quantitatively is non-trivial due to the subjective nature of this task. Inspired by~\cite{dai2017towards} that uses multiple evaluators that are trained discriminatively for assessing the quality of generated image captions, we also prepare multiple aesthetic evaluators for evaluation.
These evaluators are either elaboratively trained, \ie, AVA-Net and CUHK-Net (which have balanced accuracy of 89.1\% and 94.3\% on the CUHK-PQ dataset, respectively); or publicly available, \ie, the RANK~\cite{kong2016photo} and DAN~\cite{deng2017image}.

The results on test images are summarized respectively in Table~\ref{tab:quantitative_aesthetics1} and Table~\ref{tab:quantitative_aesthetics2}.
A higher aesthetic score suggests better aesthetic quality. 
Compared with the fully-supervised DSLR model~\cite{ignatov2017dslr}, our weakly-supervised EnhanceGAN has received consistently better quantitative ratings in terms of aesthetic scores by all aesthetic evaluators (see Table~\ref{tab:quantitative_aesthetics1}). This result is also consistent with the user study described next. Note that the aesthetic-based evaluation may not be fair to DSLR model~\cite{ignatov2017dslr} since their objective function is not to optimize aesthetic quality but focuses more on improving image sharpness, texture details and small color variations.
By contrast, our EnhanceGAN renders perceptually better aesthetic-driven color quality (see Fig.~\ref{fig:BigFig}). 
\color{black}
We further evaluate our EnhanceGAN on MIT-Adobe FiveK Dataset. We observe that our EnhanceGAN produces competitive results compared to the commercial baseline of Photoshop-Auto\footnote{Photoshop ``autoTone+autoContrast+autoColor''} and \textit{Top2-Expert} (see Table~\ref{tab:quantitative_aesthetics2}).
\color{black}
We also evaluate our EnhanceGAN on the 5-image evaluation set in~\cite{lee2016automatic}. As shown in Fig.~\ref{fig:linzhe}, our EnhanceGAN produces consistently more natural color enhancement results while the exemplar-based method~\cite{lee2016automatic} features a more aggressive change in image styles.
We also show an ablation test on different losses in Table~\ref{tab:quantitative_aesthetics1}. We use the piecewise color enhancer in this test. In general, all losses contribute to the performance of EnhanceGAN, with $L_{gan}$ playing the dominant role.
It is noteworthy that the deep filtering-based operator of EnhanceGAN performs much better than the random weights baseline, as shown in Table~\ref{tab:quantitative_aesthetics1} and Fig.~\ref{fig:BigFig}. The results suggest that EnhanceGAN learns meaningful parameters driven by image aesthetics.
%
\begin{figure}[t]
\begin{center}
\includegraphics[width=\linewidth]{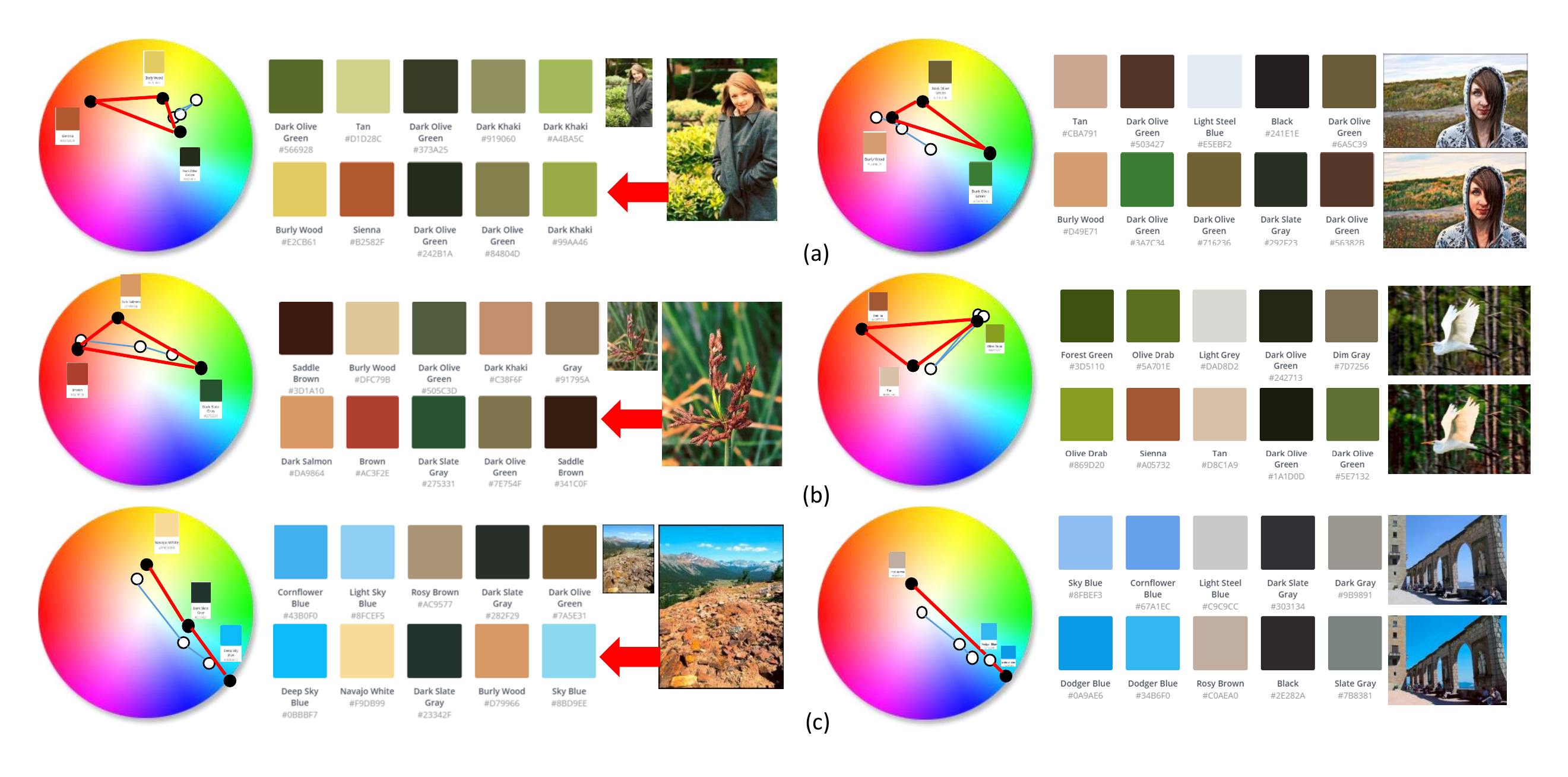}
\end{center}
\vskip -0.4cm
\caption[]{Examples of image color palettes. (a) first row: Human, enhanced outputs show \textit{Split-complimentary} harmony patterns; (b) second row: Nature, enhanced outputs show \textit{Split-complimentary} or \textit{Triadic} harmony patterns (c) third row: Skyscape, enhanced outputs show \textit{Complimentary} harmony patterns. (Best viewed in color)}
\label{fig:color_harmony}
\end{figure}
\begin{figure}[t]
\begin{center}
\includegraphics[width=\linewidth]{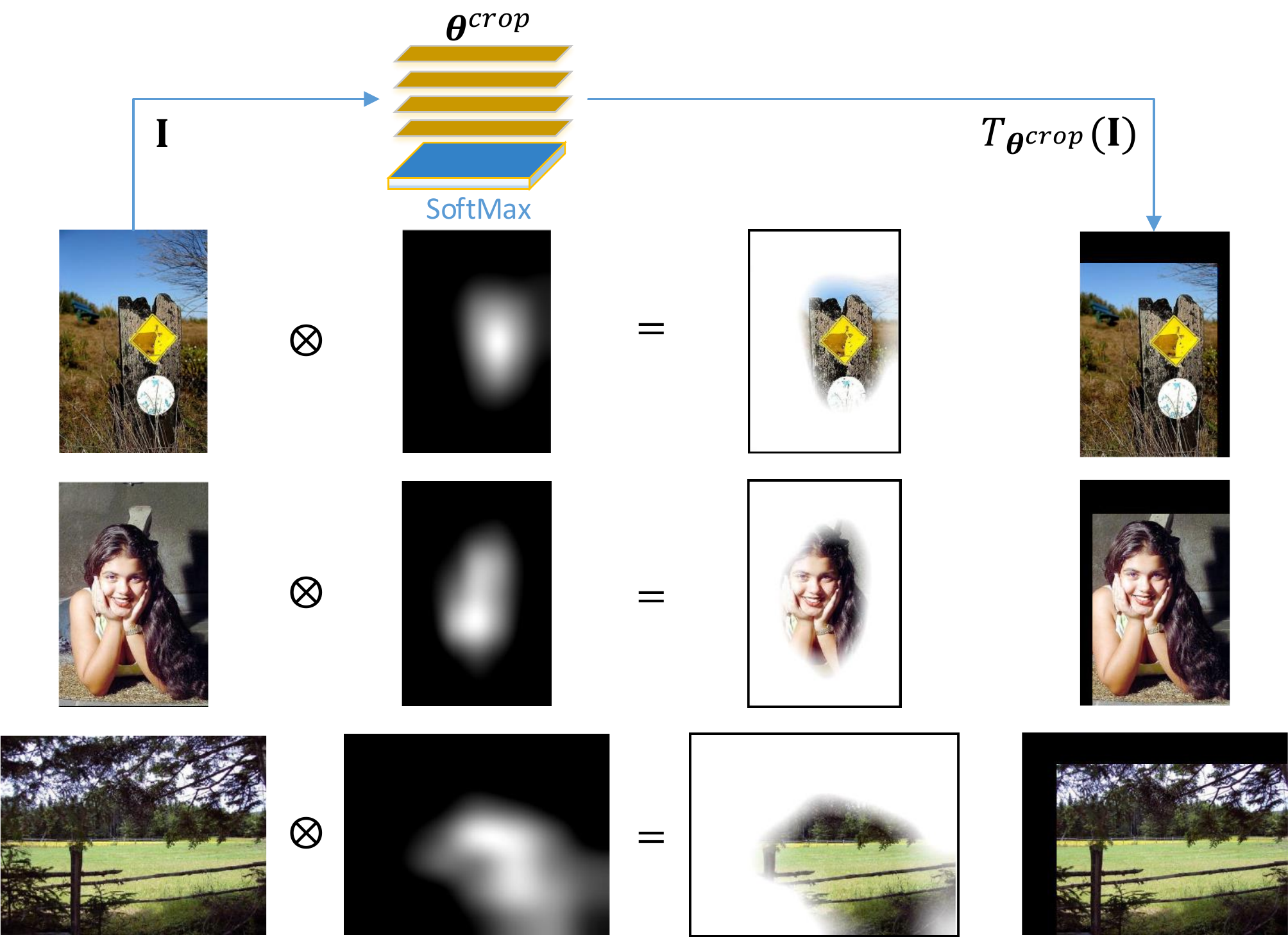}
\end{center}
\vskip -0.4cm
\caption{The focus of attention as revealed by overlaying the softmax feature map (as in Fig.~\ref{fig:architecture}$b$) onto the input image. (Best viewed in color.)}
\label{fig:cropping_response}
\vskip -0.4cm
\end{figure}
\vspace{0.1cm}
\\
\noindent\textbf{Automatic Image Cropping:} We conduct an additional experiment to demonstrate the extensibility of the proposed method.
We quantitatively evaluate the performance of our EnhanceGAN for image cropping on the popular CUHK Image Cropping Dataset~\cite{yan2013learning}, which contains 950 images and 3 sets of cropping groundtruth by 3 different annotators. 
We perform a 5-fold cross-validation test for all the supervised baselines. Note that these baselines are fine-tuned using the CUHK Image Cropping Dataset, while EnhanceGAN is \textit{NOT} finetuned with any groundtruth cropping labels. Despite the weakly-supervised nature of our approach, EnhanceGAN achieves competitive performance and even surpasses some methods with full supervision\footnote{\color{black}The non-parametric retrieval-based method by Chen \etal~\cite{chen2017learning} included groundtruth crops in their model candidates for crop selection/evaluation, which is not comparable to the reported benchmarks as in~\cite{yan2013learning,chen2017quantitative,deng2017image}.\color{black}}, as shown in Table~\ref{tab:cropping-results-cuhk}. 
\color{black} Still, it is unfair to directly compare EnhanceGAN to the fully-supervised cropping methods. Following the learning scheme in Deng et al.~\cite{deng2017image}, we show that our EnhanceGAN can be further finetuned (FT) towards state-of-the-art performance when groundtruth labels are available (see Table~\ref{tab:cropping-results-cuhk}).
\color{black}
We also observe that the cropping operator of EnhanceGAN has learned to be attentive to specific regions of the input that are relevant to the image content, hence producing reasonable crop-coordinate candidates resided on the corresponding neurons of the feature maps (see Fig.~\ref{fig:cropping_response}). 
%
%
%
\subsection{Color Harmony}
We spotted interesting patterns regarding the enhanced outputs by our proposed model. As shown in Figure~\ref{fig:color_harmony}, EnhanceGAN learns to produce color palettes that approximate certain color harmony schemes~\cite{cohen2006color}, such as \textit{Complementary}, \textit{Split-complementary} and \textit{Triadic} schemes. For example, for input images from human and nature categories, the color palettes from the enhanced outputs show a sign of approximately \textit{Split-complementary} or \textit{Triadic} schemes. For images from sky-and-seascape categories, the color palettes are enhanced towards the \textit{Complementary} scheme. This further shows that the aesthetic enhancement by our weakly-supervised EnhanceGAN is consistent with harmonic color patterns.

\subsection{User Study}
\label{sec:userstudy}
The subjective nature of image enhancement evaluation also calls for a validation through human survey. For the purpose of the user study, we asked a professional editor to enhance each of the 100 images in $Val^{100}$ in \textit{Adobe Photoshop}. Image editing options including the tools ``Levels'', ``Curves'', ``Auto Tone'',  ``Auto Contrast'' and ``Auto Color'' in \textit{Adobe Photoshop} were available to the professional editor. All enhanced images were stored using sRGB JPEG-format with the highest quality (Quality = 12 in \textit{Adobe Photoshop}). We wrote a ranking software and distributed to a total of \totaluser~participants.
All participants were shown a sequence of 100 image sets, where each image set contained a random ordering of the input image, the image enhanced by the piecewise color enhancer, the image enhanced by the deep filtering-based enhancer, \color{black}the Photoshop-Auto output, \color{black} and the human edited image. Participants were instructed to rank each set by clicking the best-quality image, the excellent-quality image, the good-quality image, the average-quality image, and the poor-quality image on the screen in order. No time constraints were placed.

\begin{figure}
\begin{center}
\includegraphics[width=\linewidth]{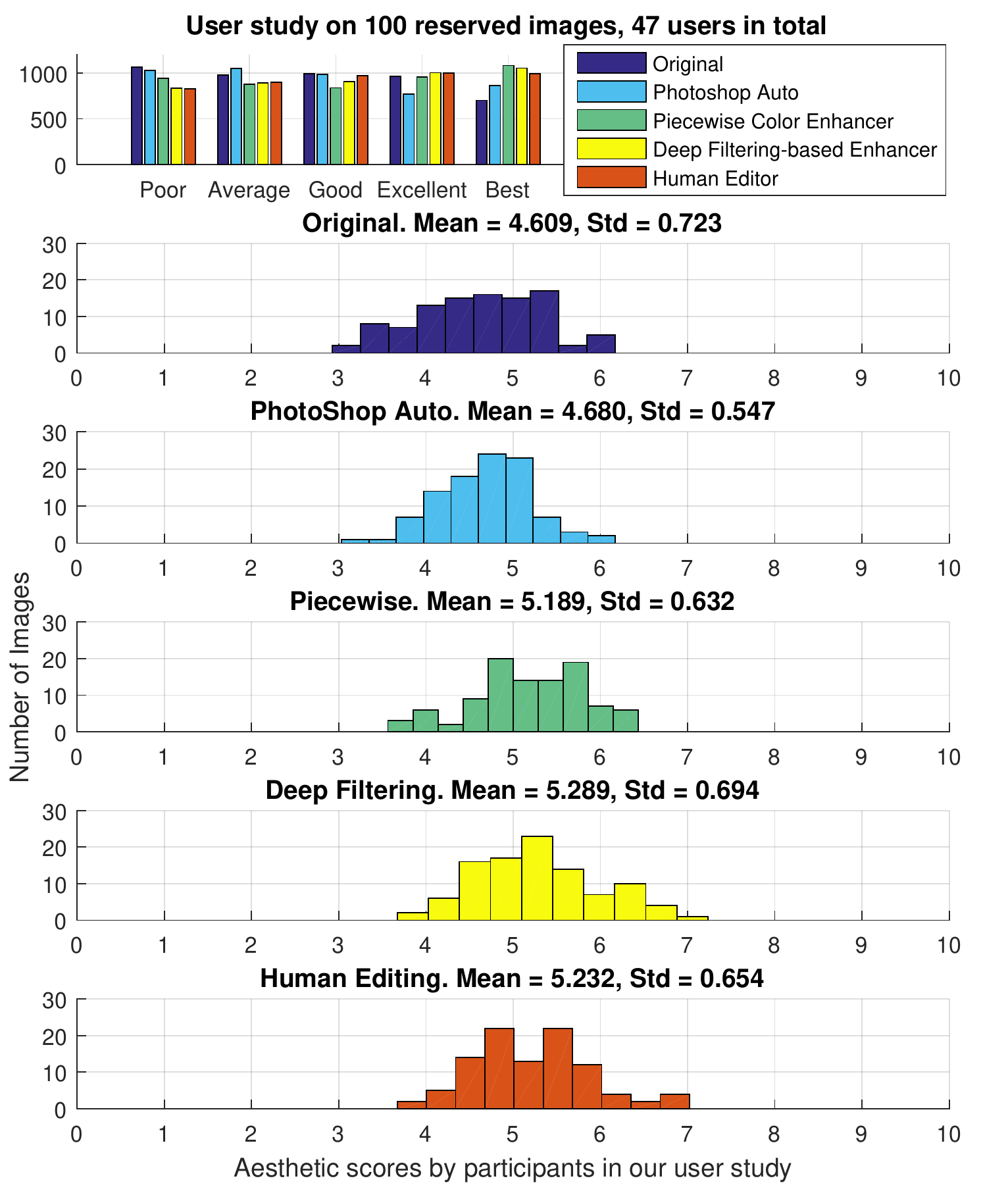}
\end{center}
\vskip -0.4cm
\caption{User study on $Val^{100}$. Our EnhanceGAN shows competitive performance as compared to human editing.}
\label{fig:userstudy}
\end{figure}

\begin{figure}[t]
\centering
\includegraphics[width=\linewidth]{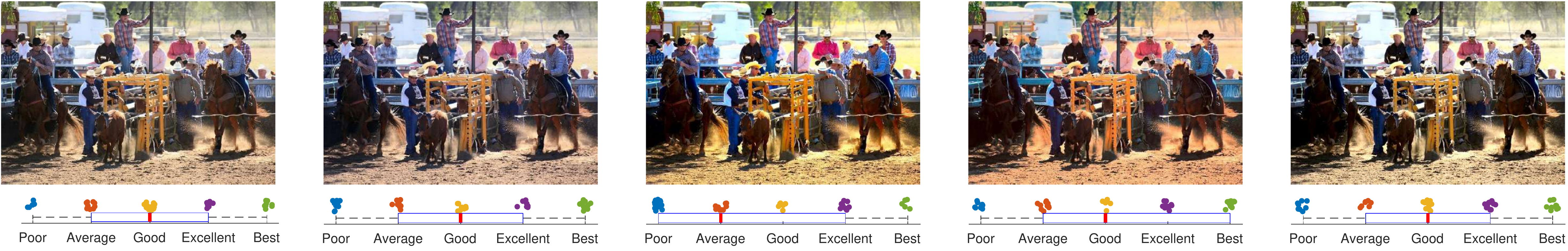}
\includegraphics[width=\linewidth]{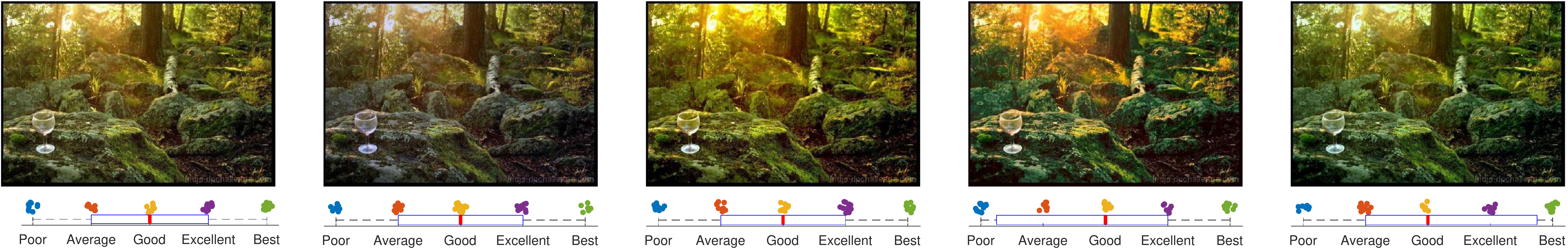}
\includegraphics[width=\linewidth]{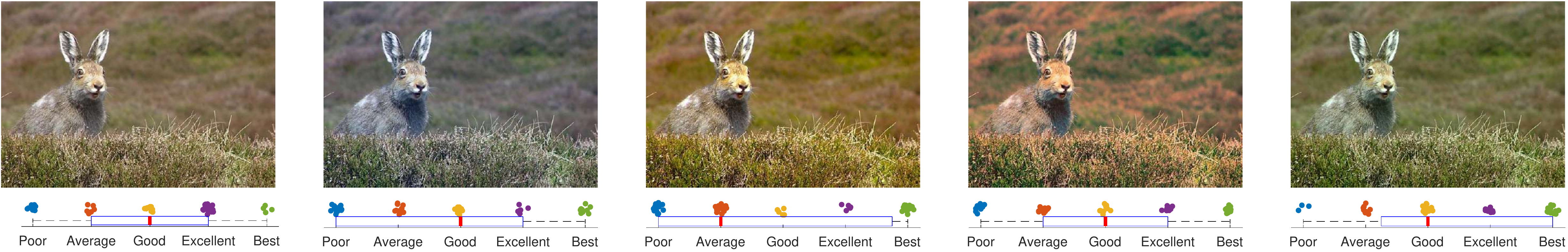}
\includegraphics[width=\linewidth]{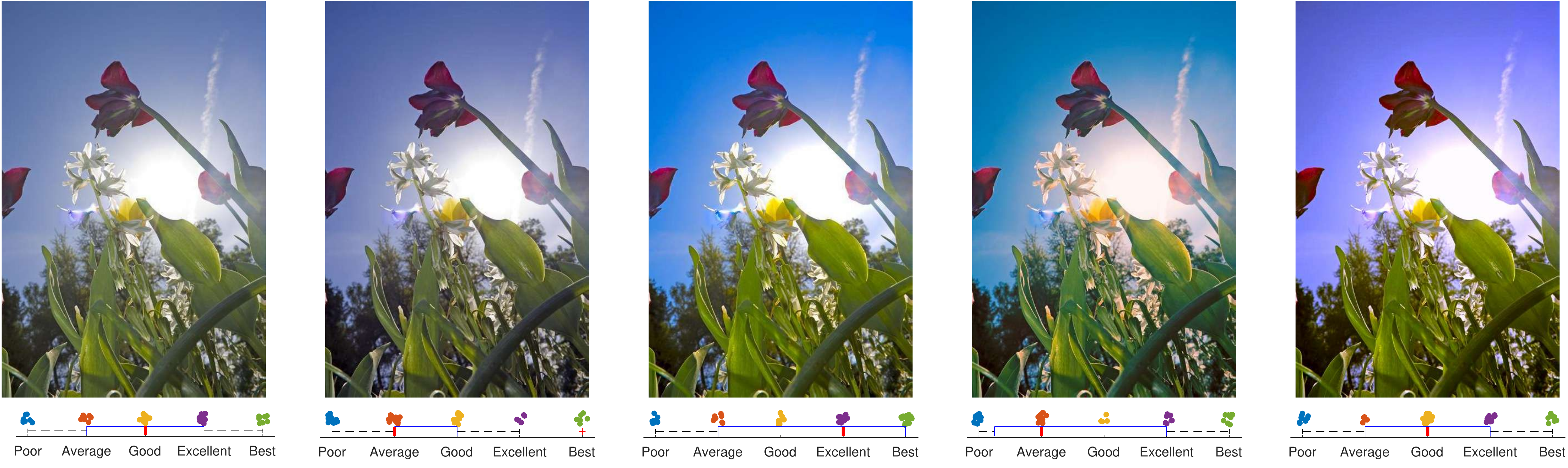}
\includegraphics[width=\linewidth]{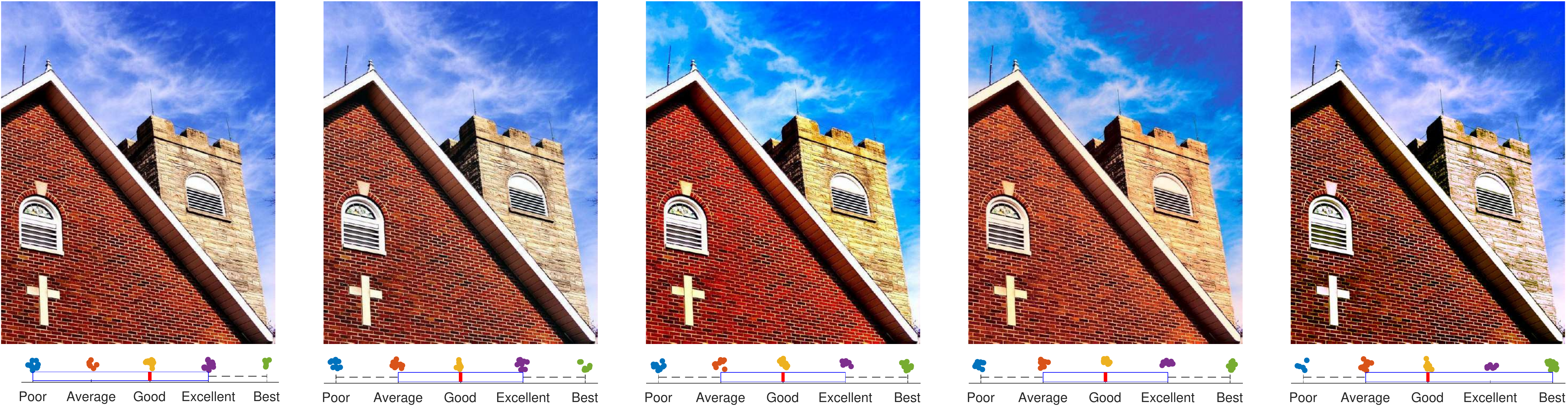}
\includegraphics[width=\linewidth]{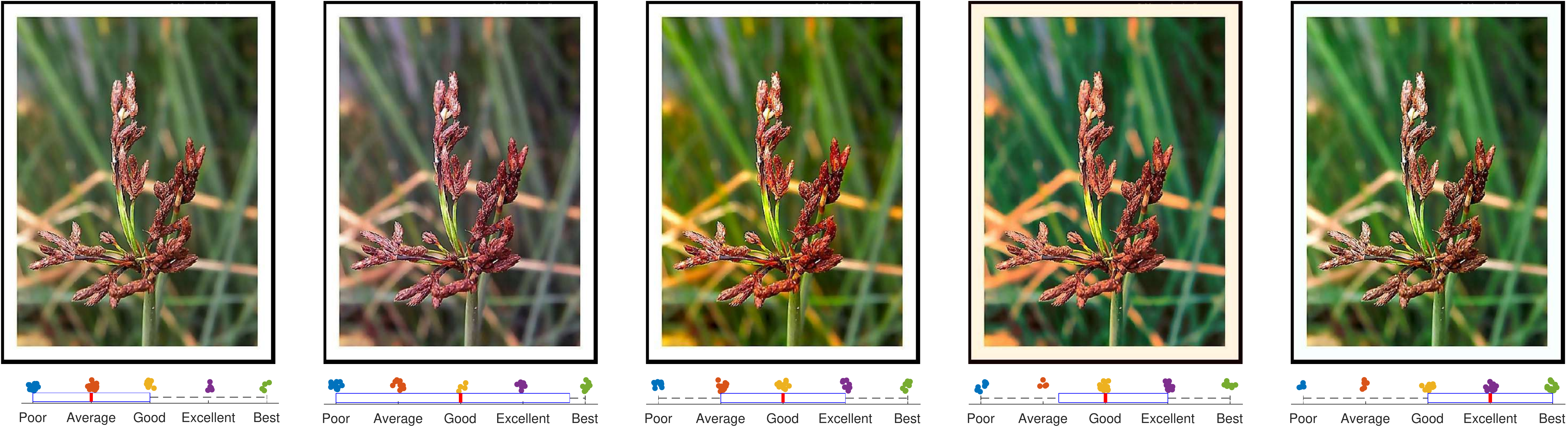}
\caption{Visual results from our user study. From left to right: (a) Original image; (b) Photoshop-Auto (c) EnhanceGAN - piecewise color enhancer; (d) EnhanceGAN - deep filtering-based enhancer; (e) Human editing. The box plot below shows the ranking for each image, and the amount of dots denotes the number of users who gives a particular rank as in $\{$Poor, Average, Good, Excellent, Best$\}$. We show the complete results in the supplementary material.}
\label{fig:visual_results}
\end{figure}

The results of our user study are shown in Fig.~\ref{fig:userstudy}. Each image in $Val^{100}$ received \totaluser~ratings, where we assign ``best quality'', \color{black}``excellent quality''\color{black},``good quality'', ``average quality'' and ``poor quality'' to aesthetic scores of $10, 7.50, 5.00, 2.50$ and $0.00$, respectively. We observe that among the images ranked as the ``best quality'' or ``excellent quality'', the majority of them are from the EnhanceGAN outputs and the human editing. Our piecewise color enhancer and deep filtering-based enhancer obtain mean aesthetic scores of $5.189$ and $5.289$, matching $5.232$ for human edited images and surpassing scores for Photoshop-Auto and the original inputs. Some of the images enhanced by EnhanceGAN receive even higher voting than the ones produced by the professional editor, as shown in Fig.~\ref{fig:visual_results}. The results demonstrate the effectiveness of the proposed EnhanceGAN for automatic image enhancement and confirm our quantitative evaluation results as in Sec.~\ref{sec:quantitative}.
\begin{table}
\centering
\caption{Quantitative evaluation on CUHK Image Cropping Dataset~\cite{yan2013learning}. The first number is average overlap ratio, higher is better. The second number (shown in parenthesis) is average boundary displacement error, lower is better~\cite{yan2013learning}. \color{black}Our EnhanceGAN is by itself competitive compared to weakly-supervised methods, and can be further finetuned (FT) towards state-of-the-art performance\color{black}.}
\color{black}
\setlength{\tabcolsep}{3.0 pt}
\begin{tabular}{lrrrrr}
Full supervision    & \small{Photographer1} &  \small{Photographer2} & \small{Photographer3} \\ \hline
Park \etal \cite{park2012modeling}                           & 0.603 (0.106)      &  0.582 (0.113) & 0.609 (0.110)         \\
Yan \etal \cite{yan2013learning}                              & 0.749 (0.067)        &  0.729 (0.072) & 0.732 (0.072)         \\
\color{black} A2-RL \cite{li2017a2}                                            & 0.793 (0.054)           & 0.791 (0.055) & 0.783 (0.055) \color{black} \\
Deng \etal \cite{deng2017image}                            & 0.806 (0.031)          & 0.775 (0.038) & 0.773 (0.038)  \\
EnhanceGAN (FT)                                                 & $\mathbf{0.828~(0.043)}$          & $\mathbf{0.805 ~(0.050)}$ & $\mathbf{0.798~(0.051)}$  \\
& & & \\
Weak supervision     & & & \\ \hline
Chen \etal    \cite{chen2017quantitative}                 & 0.664 (0.092)      &  0.656 (0.095) & 0.644 (0.099)         \\
EnhanceGAN                                                 & $\mathbf{0.715~(0.077)}$        &  $\mathbf{0.701 ~(0.081)}$ & $\mathbf{0.702~(0.080)}$         \\
\end{tabular}
\label{tab:cropping-results-cuhk}
\vskip -0.4cm
\end{table}

\section{Conclusion} 
We have introduced EnhanceGAN for automatic image enhancement. Unlike most existing approaches that require well-aligned and paired images for training, EnhanceGAN only requires weak supervision in the form of binary label on aesthetic quality. We have demonstrated its capability in learning different enhancement operators in an aesthetic-driven manner. 
EnhanceGAN is fully-differentiable and can be trained end-to-end.
We have quantitatively evaluated the performance of EnhanceGAN, and through a user study, we have shown that the high-quality results produced by EnhanceGAN are on par with professional editing.

\bibliographystyle{ACM-Reference-Format}
\bibliography{danny_cvpr18}


\begin{thebibliography}{45}


\ifx \showCODEN    \undefined \def \showCODEN     #1{\unskip}     \fi
\ifx \showDOI      \undefined \def \showDOI       #1{#1}\fi
\ifx \showISBNx    \undefined \def \showISBNx     #1{\unskip}     \fi
\ifx \showISBNxiii \undefined \def \showISBNxiii  #1{\unskip}     \fi
\ifx \showISSN     \undefined \def \showISSN      #1{\unskip}     \fi
\ifx \showLCCN     \undefined \def \showLCCN      #1{\unskip}     \fi
\ifx \shownote     \undefined \def \shownote      #1{#1}          \fi
\ifx \showarticletitle \undefined \def \showarticletitle #1{#1}   \fi
\ifx \showURL      \undefined \def \showURL       {\relax}        \fi
\providecommand\bibfield[2]{#2}
\providecommand\bibinfo[2]{#2}
\providecommand\natexlab[1]{#1}
\providecommand\showeprint[2][]{arXiv:#2}

\bibitem[\protect\citeauthoryear{Arjovsky, Chintala, and Bottou}{Arjovsky
  et~al\mbox{.}}{2017}]%
        {arjovsky2017wasserstein}
\bibfield{author}{\bibinfo{person}{Martin Arjovsky}, \bibinfo{person}{Soumith
  Chintala}, {and} \bibinfo{person}{L{\'e}on Bottou}.}
  \bibinfo{year}{2017}\natexlab{}.
\newblock \showarticletitle{Wasserstein gan}.
\newblock \bibinfo{journal}{\emph{arXiv:1701.07875}} (\bibinfo{year}{2017}).
\newblock


\bibitem[\protect\citeauthoryear{Bhattacharya, Sukthankar, and
  Shah}{Bhattacharya et~al\mbox{.}}{2010}]%
        {bhattacharya2010framework}
\bibfield{author}{\bibinfo{person}{Subhabrata Bhattacharya},
  \bibinfo{person}{Rahul Sukthankar}, {and} \bibinfo{person}{Mubarak Shah}.}
  \bibinfo{year}{2010}\natexlab{}.
\newblock \showarticletitle{A framework for photo-quality assessment and
  enhancement based on visual aesthetics}. In
  \bibinfo{booktitle}{\emph{ACMMM}}.
\newblock


\bibitem[\protect\citeauthoryear{Bychkovsky, Paris, Chan, and
  Durand}{Bychkovsky et~al\mbox{.}}{2011}]%
        {bychkovsky2011learning}
\bibfield{author}{\bibinfo{person}{Vladimir Bychkovsky},
  \bibinfo{person}{Sylvain Paris}, \bibinfo{person}{Eric Chan}, {and}
  \bibinfo{person}{Fr{\'e}do Durand}.} \bibinfo{year}{2011}\natexlab{}.
\newblock \showarticletitle{Learning Photographic Global Tonal Adjustment with
  a Database of Input / Output Image Pairs}. In
  \bibinfo{booktitle}{\emph{CVPR}}.
\newblock


\bibitem[\protect\citeauthoryear{Chen, Bai, Liang, and Li}{Chen
  et~al\mbox{.}}{2016}]%
        {chen2016automatic}
\bibfield{author}{\bibinfo{person}{Jiansheng Chen}, \bibinfo{person}{Gaocheng
  Bai}, \bibinfo{person}{Shaoheng Liang}, {and} \bibinfo{person}{Zhengqin Li}.}
  \bibinfo{year}{2016}\natexlab{}.
\newblock \showarticletitle{Automatic image cropping: A computational
  complexity study}. In \bibinfo{booktitle}{\emph{CVPR}}.
\newblock


\bibitem[\protect\citeauthoryear{Chen, Huang, Chang, Tsai, Chen, and Chen}{Chen
  et~al\mbox{.}}{2017a}]%
        {chen2017quantitative}
\bibfield{author}{\bibinfo{person}{Yi-Ling Chen}, \bibinfo{person}{Tzu-Wei
  Huang}, \bibinfo{person}{Kai-Han Chang}, \bibinfo{person}{Yu-Chen Tsai},
  \bibinfo{person}{Hwann-Tzong Chen}, {and} \bibinfo{person}{Bing-Yu Chen}.}
  \bibinfo{year}{2017}\natexlab{a}.
\newblock \showarticletitle{Quantitative Analysis of Automatic Image Cropping
  Algorithms: A Dataset and Comparative Study}.
\newblock \bibinfo{journal}{\emph{WACV}} (\bibinfo{year}{2017}).
\newblock


\bibitem[\protect\citeauthoryear{Chen, Klopp, Sun, Chien, and Ma}{Chen
  et~al\mbox{.}}{2017b}]%
        {chen2017learning}
\bibfield{author}{\bibinfo{person}{Yi-Ling Chen}, \bibinfo{person}{Jan Klopp},
  \bibinfo{person}{Min Sun}, \bibinfo{person}{Shao-Yi Chien}, {and}
  \bibinfo{person}{Kwan-Liu Ma}.} \bibinfo{year}{2017}\natexlab{b}.
\newblock \showarticletitle{Learning to Compose with Professional Photographs
  on the Web}. In \bibinfo{booktitle}{\emph{ACMMM}}.
\newblock


\bibitem[\protect\citeauthoryear{Choi and Kim}{Choi and Kim}{2016}]%
        {choi2016object}
\bibfield{author}{\bibinfo{person}{Jiwon Choi} {and} \bibinfo{person}{Changick
  Kim}.} \bibinfo{year}{2016}\natexlab{}.
\newblock \showarticletitle{Object-aware image thumbnailing using image
  classification and enhanced detection of ROI}.
\newblock \bibinfo{journal}{\emph{Multimedia Tools and Applications}}
  \bibinfo{volume}{75}, \bibinfo{number}{23} (\bibinfo{year}{2016}).
\newblock


\bibitem[\protect\citeauthoryear{Dai, Lin, Urtasun, and Fidler}{Dai
  et~al\mbox{.}}{2017}]%
        {dai2017towards}
\bibfield{author}{\bibinfo{person}{Bo Dai}, \bibinfo{person}{Dahua Lin},
  \bibinfo{person}{Raquel Urtasun}, {and} \bibinfo{person}{Sanja Fidler}.}
  \bibinfo{year}{2017}\natexlab{}.
\newblock \showarticletitle{Towards Diverse and Natural Image Descriptions via
  a Conditional GAN}. In \bibinfo{booktitle}{\emph{ICCV}}.
\newblock


\bibitem[\protect\citeauthoryear{Deng, Loy, and Tang}{Deng
  et~al\mbox{.}}{2017}]%
        {deng2017image}
\bibfield{author}{\bibinfo{person}{Yubin Deng}, \bibinfo{person}{Chen~Change
  Loy}, {and} \bibinfo{person}{Xiaoou Tang}.} \bibinfo{year}{2017}\natexlab{}.
\newblock \showarticletitle{Image Aesthetic Assessment: An Experimental
  Survey}.
\newblock \bibinfo{journal}{\emph{IEEE Signal Processing Magazine}}
  \bibinfo{volume}{34}, \bibinfo{number}{4} (\bibinfo{year}{2017}),
  \bibinfo{pages}{80--106}.
\newblock


\bibitem[\protect\citeauthoryear{Dong, Loy, He, and Tang}{Dong
  et~al\mbox{.}}{2016}]%
        {dong2016image}
\bibfield{author}{\bibinfo{person}{Chao Dong}, \bibinfo{person}{Chen~Change
  Loy}, \bibinfo{person}{Kaiming He}, {and} \bibinfo{person}{Xiaoou Tang}.}
  \bibinfo{year}{2016}\natexlab{}.
\newblock \showarticletitle{Image super-resolution using deep convolutional
  networks}.
\newblock \bibinfo{journal}{\emph{TPAMI}} \bibinfo{volume}{38},
  \bibinfo{number}{2} (\bibinfo{year}{2016}).
\newblock


\bibitem[\protect\citeauthoryear{Fang, Lin, Mech, and Shen}{Fang
  et~al\mbox{.}}{2014}]%
        {fang2014automatic}
\bibfield{author}{\bibinfo{person}{Chen Fang}, \bibinfo{person}{Zhe Lin},
  \bibinfo{person}{Radomir Mech}, {and} \bibinfo{person}{Xiaohui Shen}.}
  \bibinfo{year}{2014}\natexlab{}.
\newblock \showarticletitle{Automatic image cropping using visual composition,
  boundary simplicity and content preservation models}. In
  \bibinfo{booktitle}{\emph{ACMMM}}.
\newblock


\bibitem[\protect\citeauthoryear{Faridul, Pouli, Chamaret, Stauder,
  Tr{\'e}meau, Reinhard, et~al\mbox{.}}{Faridul et~al\mbox{.}}{2014}]%
        {faridul2014survey}
\bibfield{author}{\bibinfo{person}{Hasan~Sheikh Faridul},
  \bibinfo{person}{Tania Pouli}, \bibinfo{person}{Christel Chamaret},
  \bibinfo{person}{J{\"u}rgen Stauder}, \bibinfo{person}{Alain Tr{\'e}meau},
  \bibinfo{person}{Erik Reinhard}, {et~al\mbox{.}}}
  \bibinfo{year}{2014}\natexlab{}.
\newblock \showarticletitle{A Survey of Color Mapping and its Applications.}.
  In \bibinfo{booktitle}{\emph{Eurographics (State of the Art Reports)}}.
\newblock


\bibitem[\protect\citeauthoryear{Goodfellow, Pouget-Abadie, Mirza, Xu,
  Warde-Farley, Ozair, Courville, and Bengio}{Goodfellow et~al\mbox{.}}{2014}]%
        {goodfellow2014generative}
\bibfield{author}{\bibinfo{person}{Ian Goodfellow}, \bibinfo{person}{Jean
  Pouget-Abadie}, \bibinfo{person}{Mehdi Mirza}, \bibinfo{person}{Bing Xu},
  \bibinfo{person}{David Warde-Farley}, \bibinfo{person}{Sherjil Ozair},
  \bibinfo{person}{Aaron Courville}, {and} \bibinfo{person}{Yoshua Bengio}.}
  \bibinfo{year}{2014}\natexlab{}.
\newblock \showarticletitle{Generative adversarial nets}. In
  \bibinfo{booktitle}{\emph{NIPS}}.
\newblock


\bibitem[\protect\citeauthoryear{Gupta, Johnson, Alahi, and Fei-Fei}{Gupta
  et~al\mbox{.}}{2017}]%
        {gupta2017characterizing}
\bibfield{author}{\bibinfo{person}{Agrim Gupta}, \bibinfo{person}{Justin
  Johnson}, \bibinfo{person}{Alexandre Alahi}, {and} \bibinfo{person}{Li
  Fei-Fei}.} \bibinfo{year}{2017}\natexlab{}.
\newblock \showarticletitle{Characterizing and Improving Stability in Neural
  Style Transfer}. In \bibinfo{booktitle}{\emph{ICCV}}.
\newblock


\bibitem[\protect\citeauthoryear{He, Sun, and Tang}{He et~al\mbox{.}}{2011}]%
        {he2011single}
\bibfield{author}{\bibinfo{person}{Kaiming He}, \bibinfo{person}{Jian Sun},
  {and} \bibinfo{person}{Xiaoou Tang}.} \bibinfo{year}{2011}\natexlab{}.
\newblock \showarticletitle{Single image haze removal using dark channel
  prior}.
\newblock \bibinfo{journal}{\emph{TPAMI}} \bibinfo{volume}{33},
  \bibinfo{number}{12} (\bibinfo{year}{2011}).
\newblock


\bibitem[\protect\citeauthoryear{He, Zhang, Ren, and Sun}{He
  et~al\mbox{.}}{2016}]%
        {he2016deep}
\bibfield{author}{\bibinfo{person}{Kaiming He}, \bibinfo{person}{Xiangyu
  Zhang}, \bibinfo{person}{Shaoqing Ren}, {and} \bibinfo{person}{Jian Sun}.}
  \bibinfo{year}{2016}\natexlab{}.
\newblock \showarticletitle{Deep residual learning for image recognition}. In
  \bibinfo{booktitle}{\emph{CVPR}}.
\newblock


\bibitem[\protect\citeauthoryear{Hosie-Bounar, Hart, and Geller}{Hosie-Bounar
  et~al\mbox{.}}{2011}]%
        {hosie2011new}
\bibfield{author}{\bibinfo{person}{Jane Hosie-Bounar}, \bibinfo{person}{Kelly
  Hart}, {and} \bibinfo{person}{Mitch Geller}.}
  \bibinfo{year}{2011}\natexlab{}.
\newblock \bibinfo{booktitle}{\emph{New Perspectives on Adobe Photoshop CS5,
  Comprehensive}}.
\newblock \bibinfo{publisher}{Cengage Learning}.
\newblock


\bibitem[\protect\citeauthoryear{Huang, Chen, Wang, and Lin}{Huang
  et~al\mbox{.}}{2015}]%
        {huang2015automatic}
\bibfield{author}{\bibinfo{person}{Jingwei Huang}, \bibinfo{person}{Huarong
  Chen}, \bibinfo{person}{Bin Wang}, {and} \bibinfo{person}{Stephen Lin}.}
  \bibinfo{year}{2015}\natexlab{}.
\newblock \showarticletitle{Automatic thumbnail generation based on visual
  representativeness and foreground recognizability}. In
  \bibinfo{booktitle}{\emph{ICCV}}.
\newblock


\bibitem[\protect\citeauthoryear{Huang and Belongie}{Huang and
  Belongie}{2017}]%
        {huang2017arbitrary}
\bibfield{author}{\bibinfo{person}{Xun Huang} {and} \bibinfo{person}{Serge
  Belongie}.} \bibinfo{year}{2017}\natexlab{}.
\newblock \showarticletitle{Arbitrary Style Transfer in Real-time with Adaptive
  Instance Normalization}. In \bibinfo{booktitle}{\emph{ICCV}}.
\newblock


\bibitem[\protect\citeauthoryear{Hwang, Lee, So~Kweon, and Joo~Kim}{Hwang
  et~al\mbox{.}}{2014}]%
        {hwang2014color}
\bibfield{author}{\bibinfo{person}{Youngbae Hwang}, \bibinfo{person}{Joon-Young
  Lee}, \bibinfo{person}{In So~Kweon}, {and} \bibinfo{person}{Seon Joo~Kim}.}
  \bibinfo{year}{2014}\natexlab{}.
\newblock \showarticletitle{Color transfer using probabilistic moving least
  squares}. In \bibinfo{booktitle}{\emph{CVPR}}.
\newblock


\bibitem[\protect\citeauthoryear{Ignatov, Kobyshev, Vanhoey, Timofte, and
  Van~Gool}{Ignatov et~al\mbox{.}}{2017}]%
        {ignatov2017dslr}
\bibfield{author}{\bibinfo{person}{Andrey Ignatov}, \bibinfo{person}{Nikolay
  Kobyshev}, \bibinfo{person}{Kenneth Vanhoey}, \bibinfo{person}{Radu Timofte},
  {and} \bibinfo{person}{Luc Van~Gool}.} \bibinfo{year}{2017}\natexlab{}.
\newblock \showarticletitle{{DSLR}-Quality Photos on Mobile Devices with Deep
  Convolutional Networks}. In \bibinfo{booktitle}{\emph{ICCV}}.
\newblock


\bibitem[\protect\citeauthoryear{Islam, Lai-Kuan, and Chee-Onn}{Islam
  et~al\mbox{.}}{2016}]%
        {islam2016survey}
\bibfield{author}{\bibinfo{person}{Md~Baharul Islam}, \bibinfo{person}{Wong
  Lai-Kuan}, {and} \bibinfo{person}{Wong Chee-Onn}.}
  \bibinfo{year}{2016}\natexlab{}.
\newblock \showarticletitle{A survey of aesthetics-driven image recomposition}.
\newblock \bibinfo{journal}{\emph{Multimedia Tools and Applications}}
  (\bibinfo{year}{2016}).
\newblock


\bibitem[\protect\citeauthoryear{Jaderberg, Simonyan, Zisserman,
  et~al\mbox{.}}{Jaderberg et~al\mbox{.}}{2015}]%
        {jaderberg2015spatial}
\bibfield{author}{\bibinfo{person}{Max Jaderberg}, \bibinfo{person}{Karen
  Simonyan}, \bibinfo{person}{Andrew Zisserman}, {et~al\mbox{.}}}
  \bibinfo{year}{2015}\natexlab{}.
\newblock \showarticletitle{Spatial transformer networks}. In
  \bibinfo{booktitle}{\emph{NIPS}}.
\newblock


\bibitem[\protect\citeauthoryear{Jaiswal and Meghrajani}{Jaiswal and
  Meghrajani}{2015}]%
        {jaiswal2015saliency}
\bibfield{author}{\bibinfo{person}{Nehal Jaiswal} {and}
  \bibinfo{person}{Yogesh~K Meghrajani}.} \bibinfo{year}{2015}\natexlab{}.
\newblock \showarticletitle{Saliency based automatic image cropping using
  support vector machine classifier}. In
  \bibinfo{booktitle}{\emph{International Conference on Innovations in
  Information, Embedded and Communication Systems}}.
\newblock


\bibitem[\protect\citeauthoryear{Johnson, Alahi, and Fei-Fei}{Johnson
  et~al\mbox{.}}{2016}]%
        {johnson2016perceptual}
\bibfield{author}{\bibinfo{person}{Justin Johnson}, \bibinfo{person}{Alexandre
  Alahi}, {and} \bibinfo{person}{Li Fei-Fei}.} \bibinfo{year}{2016}\natexlab{}.
\newblock \showarticletitle{Perceptual losses for real-time style transfer and
  super-resolution}. In \bibinfo{booktitle}{\emph{ECCV}}. Springer.
\newblock


\bibitem[\protect\citeauthoryear{Kong, Shen, Lin, Mech, and Fowlkes}{Kong
  et~al\mbox{.}}{2016}]%
        {kong2016photo}
\bibfield{author}{\bibinfo{person}{Shu Kong}, \bibinfo{person}{Xiaohui Shen},
  \bibinfo{person}{Zhe Lin}, \bibinfo{person}{Radomir Mech}, {and}
  \bibinfo{person}{Charless Fowlkes}.} \bibinfo{year}{2016}\natexlab{}.
\newblock \showarticletitle{Photo aesthetics ranking network with attributes
  and content adaptation}. In \bibinfo{booktitle}{\emph{ECCV}}. Springer.
\newblock


\bibitem[\protect\citeauthoryear{Lee, Sunkavalli, Lin, Shen, and So~Kweon}{Lee
  et~al\mbox{.}}{2016}]%
        {lee2016automatic}
\bibfield{author}{\bibinfo{person}{Joon-Young Lee}, \bibinfo{person}{Kalyan
  Sunkavalli}, \bibinfo{person}{Zhe Lin}, \bibinfo{person}{Xiaohui Shen}, {and}
  \bibinfo{person}{In So~Kweon}.} \bibinfo{year}{2016}\natexlab{}.
\newblock \showarticletitle{Automatic content-aware color and tone
  stylization}. In \bibinfo{booktitle}{\emph{CVPR}}.
\newblock


\bibitem[\protect\citeauthoryear{Li, Wu, Zhang, and Huang}{Li
  et~al\mbox{.}}{2017}]%
        {li2017a2}
\bibfield{author}{\bibinfo{person}{Debang Li}, \bibinfo{person}{Huikai Wu},
  \bibinfo{person}{Junge Zhang}, {and} \bibinfo{person}{Kaiqi Huang}.}
  \bibinfo{year}{2017}\natexlab{}.
\newblock \showarticletitle{A2-RL: Aesthetics Aware Reinforcement Learning for
  Automatic Image Cropping}.
\newblock \bibinfo{journal}{\emph{arXiv preprint arXiv:1709.04595}}
  (\bibinfo{year}{2017}).
\newblock


\bibitem[\protect\citeauthoryear{Lu, Lin, Jin, Yang, and Wang}{Lu
  et~al\mbox{.}}{2014}]%
        {lu2014rapid}
\bibfield{author}{\bibinfo{person}{Xin Lu}, \bibinfo{person}{Zhe Lin},
  \bibinfo{person}{Hailin Jin}, \bibinfo{person}{Jianchao Yang}, {and}
  \bibinfo{person}{James~Z Wang}.} \bibinfo{year}{2014}\natexlab{}.
\newblock \showarticletitle{Rapid: Rating pictorial aesthetics using deep
  learning}. In \bibinfo{booktitle}{\emph{ACMMM}}.
\newblock


\bibitem[\protect\citeauthoryear{Ma, Liu, and Chen}{Ma et~al\mbox{.}}{2017}]%
        {ma2017lamp}
\bibfield{author}{\bibinfo{person}{Shuang Ma}, \bibinfo{person}{Jing Liu},
  {and} \bibinfo{person}{Chang~Wen Chen}.} \bibinfo{year}{2017}\natexlab{}.
\newblock \showarticletitle{A-{L}amp: Adaptive Layout-Aware Multi-Patch Deep
  Convolutional Neural Network for Photo Aesthetic Assessment}. In
  \bibinfo{booktitle}{\emph{ICCV}}.
\newblock


\bibitem[\protect\citeauthoryear{Murray, Marchesotti, and Perronnin}{Murray
  et~al\mbox{.}}{2012}]%
        {murray2012ava}
\bibfield{author}{\bibinfo{person}{Naila Murray}, \bibinfo{person}{Luca
  Marchesotti}, {and} \bibinfo{person}{Florent Perronnin}.}
  \bibinfo{year}{2012}\natexlab{}.
\newblock \showarticletitle{{AVA}: A large-scale database for aesthetic visual
  analysis}. In \bibinfo{booktitle}{\emph{CVPR}}.
\newblock


\bibitem[\protect\citeauthoryear{Park, Lee, Tai, and Kweon}{Park
  et~al\mbox{.}}{2012}]%
        {park2012modeling}
\bibfield{author}{\bibinfo{person}{Jaesik Park}, \bibinfo{person}{Joon-Young
  Lee}, \bibinfo{person}{Yu-Wing Tai}, {and} \bibinfo{person}{In~So Kweon}.}
  \bibinfo{year}{2012}\natexlab{}.
\newblock \showarticletitle{Modeling photo composition and its application to
  photo re-arrangement}. In \bibinfo{booktitle}{\emph{ICIP}}.
\newblock


\bibitem[\protect\citeauthoryear{Pelli and Zhang}{Pelli and Zhang}{1991}]%
        {pelli1991accurate}
\bibfield{author}{\bibinfo{person}{Denis~G Pelli} {and} \bibinfo{person}{Lan
  Zhang}.} \bibinfo{year}{1991}\natexlab{}.
\newblock \showarticletitle{Accurate control of contrast on microcomputer
  displays}.
\newblock \bibinfo{journal}{\emph{Vision research}} (\bibinfo{year}{1991}).
\newblock


\bibitem[\protect\citeauthoryear{Reinhard, Adhikhmin, Gooch, and
  Shirley}{Reinhard et~al\mbox{.}}{2001}]%
        {reinhard2001color}
\bibfield{author}{\bibinfo{person}{Erik Reinhard}, \bibinfo{person}{Michael
  Adhikhmin}, \bibinfo{person}{Bruce Gooch}, {and} \bibinfo{person}{Peter
  Shirley}.} \bibinfo{year}{2001}\natexlab{}.
\newblock \showarticletitle{Color transfer between images}.
\newblock \bibinfo{journal}{\emph{IEEE Computer Graphics and Applications}}
  \bibinfo{volume}{21}, \bibinfo{number}{5} (\bibinfo{year}{2001}).
\newblock


\bibitem[\protect\citeauthoryear{Simonyan and Zisserman}{Simonyan and
  Zisserman}{2014}]%
        {simonyan2014very}
\bibfield{author}{\bibinfo{person}{Karen Simonyan} {and}
  \bibinfo{person}{Andrew Zisserman}.} \bibinfo{year}{2014}\natexlab{}.
\newblock \showarticletitle{Very deep convolutional networks for large-scale
  image recognition}.
\newblock \bibinfo{journal}{\emph{arXiv:1409.1556}} (\bibinfo{year}{2014}).
\newblock


\bibitem[\protect\citeauthoryear{Sun, Chao, Kuo, and Hsu}{Sun
  et~al\mbox{.}}{2016}]%
        {sun2016photo}
\bibfield{author}{\bibinfo{person}{Wei-Tse Sun}, \bibinfo{person}{Ting-Hsuan
  Chao}, \bibinfo{person}{Yin-Hsi Kuo}, {and} \bibinfo{person}{Winston~H Hsu}.}
  \bibinfo{year}{2016}\natexlab{}.
\newblock \showarticletitle{Photo Filter Recommendation by Category-Aware
  Aesthetic Learning}.
\newblock \bibinfo{journal}{\emph{arXiv:1608.05339}} (\bibinfo{year}{2016}).
\newblock


\bibitem[\protect\citeauthoryear{Tang, Luo, and Wang}{Tang
  et~al\mbox{.}}{2013}]%
        {tang2013content}
\bibfield{author}{\bibinfo{person}{Xiaoou Tang}, \bibinfo{person}{Wei Luo},
  {and} \bibinfo{person}{Xiaogang Wang}.} \bibinfo{year}{2013}\natexlab{}.
\newblock \showarticletitle{Content-based photo quality assessment}.
\newblock \bibinfo{journal}{\emph{TMM}} \bibinfo{volume}{15},
  \bibinfo{number}{8} (\bibinfo{year}{2013}).
\newblock


\bibitem[\protect\citeauthoryear{Tieleman and Hinton}{Tieleman and
  Hinton}{2012}]%
        {tieleman2012lecture}
\bibfield{author}{\bibinfo{person}{Tijmen Tieleman} {and}
  \bibinfo{person}{Geoffrey Hinton}.} \bibinfo{year}{2012}\natexlab{}.
\newblock \showarticletitle{Lecture 6.5-rmsprop: Divide the gradient by a
  running average of its recent magnitude}.
\newblock \bibinfo{journal}{\emph{COURSERA: Neural networks for machine
  learning}} \bibinfo{volume}{4}, \bibinfo{number}{2} (\bibinfo{year}{2012}).
\newblock


\bibitem[\protect\citeauthoryear{Ulyanov, Vedaldi, and Lempitsky}{Ulyanov
  et~al\mbox{.}}{2016}]%
        {ulyanov2016instance}
\bibfield{author}{\bibinfo{person}{Dmitry Ulyanov}, \bibinfo{person}{Andrea
  Vedaldi}, {and} \bibinfo{person}{Victor Lempitsky}.}
  \bibinfo{year}{2016}\natexlab{}.
\newblock \showarticletitle{Instance Normalization: The Missing Ingredient for
  Fast Stylization}.
\newblock \bibinfo{journal}{\emph{arXiv:1607.08022}} (\bibinfo{year}{2016}).
\newblock


\bibitem[\protect\citeauthoryear{Wang, Xiong, Lin, and Van~Gool}{Wang
  et~al\mbox{.}}{2017}]%
        {wang2017untrimmednets}
\bibfield{author}{\bibinfo{person}{Limin Wang}, \bibinfo{person}{Yuanjun
  Xiong}, \bibinfo{person}{Dahua Lin}, {and} \bibinfo{person}{Luc Van~Gool}.}
  \bibinfo{year}{2017}\natexlab{}.
\newblock \showarticletitle{UntrimmedNets for Weakly Supervised Action
  Recognition and Detection}. In \bibinfo{booktitle}{\emph{CVPR}}.
\newblock


\bibitem[\protect\citeauthoryear{Wang, Liu, Chang, Ling, Yang, and Huang}{Wang
  et~al\mbox{.}}{2016}]%
        {wang2016d3}
\bibfield{author}{\bibinfo{person}{Zhangyang Wang}, \bibinfo{person}{Ding Liu},
  \bibinfo{person}{Shiyu Chang}, \bibinfo{person}{Qing Ling},
  \bibinfo{person}{Yingzhen Yang}, {and} \bibinfo{person}{Thomas~S Huang}.}
  \bibinfo{year}{2016}\natexlab{}.
\newblock \showarticletitle{D3: Deep dual-domain based fast restoration of
  JPEG-compressed images}. In \bibinfo{booktitle}{\emph{CVPR}}.
\newblock


\bibitem[\protect\citeauthoryear{Xu, Ba, Kiros, Cho, Courville, Salakhutdinov,
  Zemel, and Bengio}{Xu et~al\mbox{.}}{2015}]%
        {xu2015show}
\bibfield{author}{\bibinfo{person}{Kelvin Xu}, \bibinfo{person}{Jimmy Ba},
  \bibinfo{person}{Ryan Kiros}, \bibinfo{person}{Kyunghyun Cho},
  \bibinfo{person}{Aaron~C Courville}, \bibinfo{person}{Ruslan Salakhutdinov},
  \bibinfo{person}{Richard~S Zemel}, {and} \bibinfo{person}{Yoshua Bengio}.}
  \bibinfo{year}{2015}\natexlab{}.
\newblock \showarticletitle{Show, Attend and Tell: Neural Image Caption
  Generation with Visual Attention.}. In \bibinfo{booktitle}{\emph{ICML}},
  Vol.~\bibinfo{volume}{14}.
\newblock


\bibitem[\protect\citeauthoryear{Yan, Lin, Bing~Kang, and Tang}{Yan
  et~al\mbox{.}}{2013}]%
        {yan2013learning}
\bibfield{author}{\bibinfo{person}{Jianzhou Yan}, \bibinfo{person}{Stephen
  Lin}, \bibinfo{person}{Sing Bing~Kang}, {and} \bibinfo{person}{Xiaoou Tang}.}
  \bibinfo{year}{2013}\natexlab{}.
\newblock \showarticletitle{Learning the change for automatic image cropping}.
  In \bibinfo{booktitle}{\emph{CVPR}}.
\newblock


\bibitem[\protect\citeauthoryear{Yan, Lin, Bing~Kang, and Tang}{Yan
  et~al\mbox{.}}{2014}]%
        {yan2014learning}
\bibfield{author}{\bibinfo{person}{Jianzhou Yan}, \bibinfo{person}{Stephen
  Lin}, \bibinfo{person}{Sing Bing~Kang}, {and} \bibinfo{person}{Xiaoou Tang}.}
  \bibinfo{year}{2014}\natexlab{}.
\newblock \showarticletitle{A learning-to-rank approach for image color
  enhancement}. In \bibinfo{booktitle}{\emph{CVPR}}.
\newblock


\bibitem[\protect\citeauthoryear{Yan, Zhang, Wang, Paris, and Yu}{Yan
  et~al\mbox{.}}{2016}]%
        {yan2016automatic}
\bibfield{author}{\bibinfo{person}{Zhicheng Yan}, \bibinfo{person}{Hao Zhang},
  \bibinfo{person}{Baoyuan Wang}, \bibinfo{person}{Sylvain Paris}, {and}
  \bibinfo{person}{Yizhou Yu}.} \bibinfo{year}{2016}\natexlab{}.
\newblock \showarticletitle{Automatic photo adjustment using deep neural
  networks}.
\newblock \bibinfo{journal}{\emph{ToG}} \bibinfo{volume}{35},
  \bibinfo{number}{2} (\bibinfo{year}{2016}).
\newblock


\end{thebibliography}

\end{document}